\documentclass[]{article}
\usepackage[margin=0.5in]{geometry}
\usepackage{booktabs}
\usepackage{tikz}
\usepackage{hyperref}

\title{EvoAAA: An evolutionary methodology for automated \hl{neural} autoencoder architecture search}

\author{Francisco Charte%
	\thanks{Corresponding author: Francisco Charte, Computer Science Department, A3-241 Universidad de Ja\'en, Campus Las Lagunillas, 23071 Ja\'en, Spain.}\and
Antonio J. Rivera \and
Francisco Mart\'inez \and
Mar\'ia J. del Jesus}

\newcommand{\hl}[1]{{\color{black}#1}}

\begin{document}


\maketitle

\begin{abstract}
Machine learning models work better when curated features are provided to them. Feature engineering methods have been usually used as a preprocessing step to obtain or build a proper feature set. In late years, autoencoders \hl{(a specific type of symmetrical neural network)} have been widely used to perform representation learning, proving their competitiveness against classical feature engineering algorithms. The main obstacle in the use of autoencoders is finding a good architecture, a process that most experts confront manually. An automated autoencoder architecture search procedure, based on evolutionary methods, is proposed in this paper. The methodology is tested against nine heterogeneous data sets. The obtained results show the ability of this approach to find better architectures, able to concentrate most of the useful information in a minimized coding, in a reduced time.
\end{abstract}



\section{Introduction}\label{Sec.Intro}

\hl{Intelligent appliances based on machine learning systems~\cite{bishop2006pattern} can be found in many everyday tasks.} 
They are in charge of filtering spam email~\cite{guzella2009review}, detecting fraudulent transactions~\cite{bhattacharyya2011data}, recommending new products to buyers~\cite{schafer1999recommender} and many other apparently simple jobs. To do that, ML methods need to extract knowledge from raw data. The usefulness of that knowledge mostly depends on the quality of data features. The goal is to produce descriptive and/or predictive models, depending on the task at hand.

Choosing a curated set of attributes, or building a new one from the original features, tends to produce better results~\cite{Domingos2012AFU} than using raw variables. Hence the interest in feature engineering techniques in late years, including well-known preprocessing procedures~\cite{DataPreprocessing} such as feature selection~\cite{CorrelationFS,MutualInformationDS} and feature extraction~\cite{FeatureExtractionIntro}. Representation learning (REPL)~\cite{rumelhart1988learning,bengio_representation_2013} is a term tightly linked to perform feature engineering relying on deep learning techniques~\cite{lecun2015deep,goodfellow2016deep}.

Autoencoders~\cite{hinton2006reducing,vincent2008extracting,vincent2010stacked} are a modern general-purpose DL-based family of tools for facing \hl{REPL}. An AE is an unsupervised symmetric neural network~\cite{Charte:ReviewAEs} aimed to build a coding that maximizes the reconstruction of data patterns. The obtained coding can be applied to many different tasks~\cite{Charte:ShowcaseAEs}, including visualization, anomaly detection, hashing and noise removing. However, finding the proper AE architecture for each data set and function is not a trivial process. Usually, it is a challenge that experts have to deal with.

In this study EvoAAA, an automated methodology for designing AE architectures maximizing their reconstruction power when used with a specific data set, is proposed.  The job is confronted as a hard optimization problem~\cite{garey2002computers}, unfeasible to solve by means of exhaustive search. Our hypothesis is that evolutionary methods~\cite{back1993overview} would be able to find near-optimal AE architectures in a reasonable time.

To verify the competitiveness of EvoAAA a thorough experimentation, including nine data sets and \hl{five search methods, three of them based on  evolutionary optimization}, is conducted. The architectures found through this approach are way better than those retrieved by exhaustive search.

\begin{table}[t!]
\setlength{\tabcolsep}{6pt}
\renewcommand{\arraystretch}{1.25}
\hl{
\caption{Summary of terminology and acronyms}\label{Table.Terminology}
\scriptsize
\begin{tabular}{p{.2\linewidth}p{.75\linewidth}}
\toprule
\textbf{Term/Acronym} & \textbf{Description} \\
\midrule
   AE & Autoencoder \\
   ANN & Artificial Neural Network \\
   Chromosome & Codification which represents the set of parameters of an individual in the population \\
   CNN & Convolutional Neural Network  \\
   Cross-over & Operator to produce a new chromosome from two or more individuals (parents) by mixing their genes \\
   DE & Differential Evolution \\
   DL & Deep Learning \\
   Elitism & Technique which preserves the best individuals among generation in an EM \\
   EM & Evolutionary Method \\
   ES & Evolution Strategy \\
   EvoAAA & Evolutionary Methods for Automated Autoencoder Architecture search \\
   Fitness & Quality measure linked to each individual \\
   GA & Genetic Algorithm \\
   Gene & Each value in a chromosome \\
   Generation & Each one of the iterations in an evolutionary method \\
   Individual & A potential solution in the search space denoted by a chromosome and the corresponding fitness value \\
   LDA & Linear Discriminant Analysis \\
   LLE & Locally Linear Embedding \\
   ML & Machine Learning \\
   MLP & Multi-Layer Perceptron \\
   MSE & Mean Squared Error \\
   Mutation & Operator to produce a new chromosome altering genes in an existing individual \\
   PCA & Principal Component Analysis \\
   Population & Group of individuals in a generation \\
   REPL & Representation Learning \\
   SGD & Stochastic Gradient Descent \\
\bottomrule
\end{tabular}
}
\end{table}

\subsection{Problem formulation}\label{Sec.Problem}

That AEs are effective tools for \hl{REPL} is a known fact~\cite{hinton2006reducing,vincent2008extracting,vincent2010stacked}, having proved their superiority against classical feature engineering methods such as PCA, LDA, ISOMAP and LLE~\cite{pulgar2020choosing}. They are also a tool closely related to nonstandard learning~\cite{Charte:nonstandard} problems. An AE is a symmetrical ANN~\cite{Charte:ReviewAEs}, so it shares many of the characteristics of any ANN. 

Training an ANN implies adjusting the weights that connect their neurons, using the backpropagation method~\cite{hecht1992theory} and any variation of the gradient descent algorithm such as SGD~\cite{robbins1951stochastic}. The ANN architecture, i.e. number of layers, amount of units per layer, activation functions, etc., is set in advance, prior to the training process.

An inadequate ANN architecture could produce bad output results, regardless of the weights learned through the training process. If the ANN is too simple, adjusting too few parameters, it will be not able to learn. On the contrary, too complex ANNs suffer from overfitting~\cite{lawrence2000overfitting} due to existing enough parameters to memorize the whole training data. \hl{These same problems also affect AEs.} Achieving a balanced architecture, the point where the ANN extracts enough information from seen data to generalize well while processing future never seen patterns, is still an unsolved problem. As a result, dozens of papers which contribute the design of ANNs to tackle specific problems~\cite{ahmadlou2010enhanced,benamara2019real} are published every year.

The interest is in choosing a proper AE architecture to process an specific data set, so that the AE is able to learn an optimum representation of the data. Until now the design of AEs has been in charge of human experts. It is not an easy task to automate, since it is a hard combinatorial problem~\cite{garey2002computers}. In fact, finding a good architecture can be \hl{seen} as an optimization problem. This is a field where EMs~\cite{back1993overview} have shown their efficacy in the past.

Our proposal is a formulation to code any AE architecture so that it can be evolved by means of evolutionary approaches. The goal is to reduce the reconstruction error as much as possible. This way, the AE \hl{encoding} will concentrate the maximum general-purpose information, rather than a coding aimed to improve class separability or any other specific goal. \hl{Regarding evolutionary techniques, they will be used as the tool to optimize a set of parameters. The classical terminology in this field summarized in Table~\ref{Table.Terminology} along with the acronyms appearing in the text, will be used.}

\subsection{Literature review}

Feature engineering is a manual or automated task aimed to obtain a set of features better than the original one. Feature selection~\cite{DataPreprocessing} consists in choosing a subset of attributes while maintaining most useful information in the data. It can be manually performed by an expert in the field, but mostly is faced with automated methods based on feature correlation~\cite{CorrelationFS} and mutual information~\cite{MutualInformationDS}. By contrast, feature extraction methods transform the original data features to produce a new, usually reduced, set of attributes. Popular algorithms to do this are PCA~\cite{PCAHotelling} and LDA~\cite{LDA}, whose mathematical foundations are relatively easy to understand.

More advanced studies work with the hypothesis that the distribution of variables in the original data lies along a lower-dimensional space, usually known as \emph{manifold}. A manifold space works with the parameters that produce the data points in the original high-dimensional space. Finding this embedded space is the task of manifold learning~\cite{ManifoldLearning} algorithms. Unlike PCA or LDA, manifold methods apply non-linear transformations, so they fall into the non-linear dimensionality reduction~\cite{NonlinearDimRec} category.

Autoencoders, as detailed in~\cite{Charte:ReviewAEs}, are ANNs having a symmetric architecture, as shown in Figure~\ref{Fig.AEStructure}. The input and output layers have as many units as features there are in the data. Inner layers usually have fewer units, so that a more compact representation of the information hold in the data is produced. The goal is to reconstruct the input patterns into the output as faithfully as possible. 

\definecolor{light-gray}{gray}{0.95}
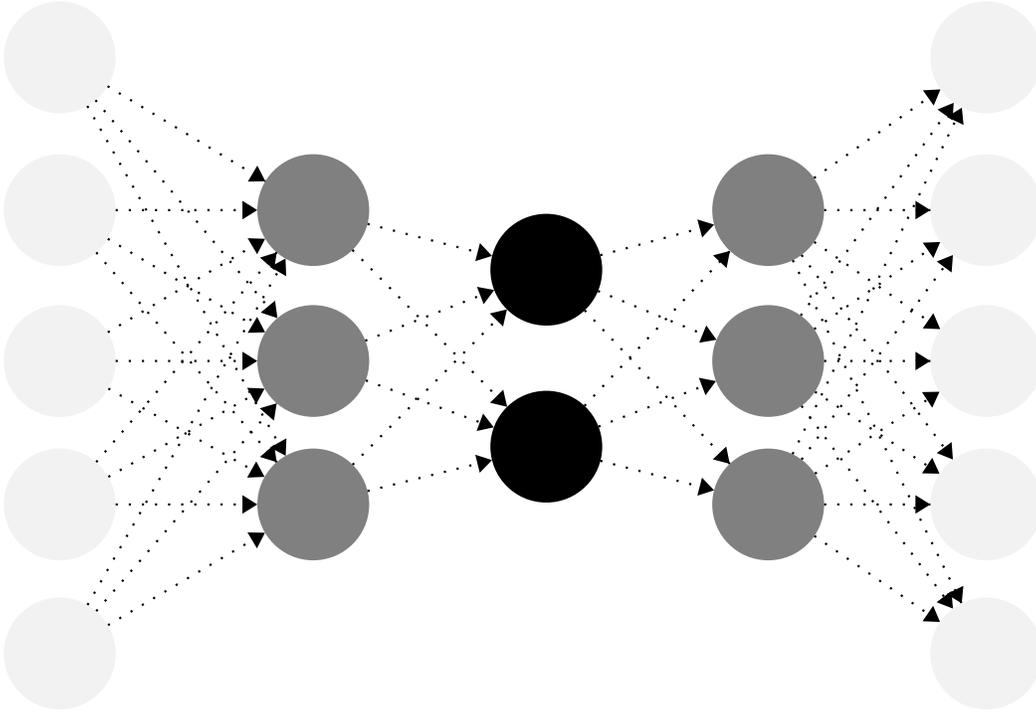
\begin{figure}[ht!]
	\centering
	\resizebox {0.75\columnwidth} {!} {
		\begin{tikzpicture}[scale=0.12]
		\tikzstyle{every node}+=[inner sep=0pt]
		\fill [light-gray,thick] (15.7,-15.6) circle (3);
		\fill [light-gray,thick] (15.7,-23.8) circle (3);
		\fill [light-gray,thick] (15.7,-31.9) circle (3);
		\fill [light-gray,thick] (15.7,-39.6) circle (3);
		\fill [light-gray,thick] (15.7,-47.6) circle (3);
		\fill [gray,thick] (29.3,-23.8) circle (3);
		\fill [gray,thick] (29.3,-31.9) circle (3);
		\fill [gray,thick] (29.3,-39.6) circle (3);
		\fill [black,thick] (41.8,-27) circle (3);
		\fill [black,thick] (41.8,-36.5) circle (3);
		\fill [gray,thick] (53.7,-23.8) circle (3);
		\fill [gray,thick] (53.7,-31.9) circle (3);
		\fill [gray,thick] (53.7,-39.6) circle (3);
		\fill [light-gray,thick] (65.4,-15.6) circle (3);
		\fill [light-gray,thick] (65.4,-23.8) circle (3);
		\fill [light-gray,thick] (65.4,-31.9) circle (3);
		\fill [light-gray,thick] (65.4,-39.6) circle (3);
		\fill [light-gray,thick] (65.4,-47.6) circle (3);
		\draw [black,dotted] (18.27,-17.15) -- (26.73,-22.25);
		\fill [black,dotted] (26.73,-22.25) -- (26.3,-21.41) -- (25.79,-22.27);
		\draw [black,dotted] (17.62,-17.9) -- (27.38,-29.6);
		\fill [black,dotted] (27.38,-29.6) -- (27.25,-28.66) -- (26.48,-29.3);
		\draw [black,dotted] (17.18,-18.21) -- (27.82,-36.99);
		\fill [black,dotted] (27.82,-36.99) -- (27.86,-36.05) -- (26.99,-36.54);
		\draw [black,dotted] (18.7,-23.8) -- (26.3,-23.8);
		\fill [black,dotted] (26.3,-23.8) -- (25.5,-23.3) -- (25.5,-24.3);
		\draw [black,dotted] (18.28,-25.34) -- (26.72,-30.36);
		\fill [black,dotted] (26.72,-30.36) -- (26.29,-29.53) -- (25.78,-30.39);
		\draw [black,dotted] (17.66,-26.07) -- (27.34,-37.33);
		\fill [black,dotted] (27.34,-37.33) -- (27.2,-36.39) -- (26.44,-37.05);
		\draw [black,dotted] (18.28,-30.36) -- (26.72,-25.34);
		\fill [black,dotted] (26.72,-25.34) -- (25.78,-25.31) -- (26.29,-26.17);
		\draw [black,dotted] (18.7,-31.9) -- (26.3,-31.9);
		\fill [black,dotted] (26.3,-31.9) -- (25.5,-31.4) -- (25.5,-32.4);
		\draw [black,dotted] (18.31,-33.38) -- (26.69,-38.12);
		\fill [black,dotted] (26.69,-38.12) -- (26.24,-37.29) -- (25.75,-38.16);
		\draw [black,dotted] (17.66,-37.33) -- (27.34,-26.07);
		\fill [black,dotted] (27.34,-26.07) -- (26.44,-26.35) -- (27.2,-27.01);
		\draw [black,dotted] (18.31,-38.12) -- (26.69,-33.38);
		\fill [black,dotted] (26.69,-33.38) -- (25.75,-33.34) -- (26.24,-34.21);
		\draw [black,dotted] (18.7,-39.6) -- (26.3,-39.6);
		\fill [black,dotted] (26.3,-39.6) -- (25.5,-39.1) -- (25.5,-40.1);
		\draw [black,dotted] (17.19,-45) -- (27.81,-26.4);
		\fill [black,dotted] (27.81,-26.4) -- (26.98,-26.85) -- (27.85,-27.35);
		\draw [black,dotted] (17.66,-45.33) -- (27.34,-34.17);
		\fill [black,dotted] (27.34,-34.17) -- (26.43,-34.44) -- (27.19,-35.1);
		\draw [black,dotted] (18.29,-46.08) -- (26.71,-41.12);
		\fill [black,dotted] (26.71,-41.12) -- (25.77,-41.1) -- (26.28,-41.96);
		\draw [black,dotted] (32.21,-24.54) -- (38.89,-26.26);
		\fill [black,dotted] (38.89,-26.26) -- (38.24,-25.57) -- (37.99,-26.54);
		\draw [black,dotted] (32.09,-30.81) -- (39.01,-28.09);
		\fill [black,dotted] (39.01,-28.09) -- (38.08,-27.92) -- (38.44,-28.85);
		\draw [black,dotted] (31.41,-37.47) -- (39.69,-29.13);
		\fill [black,dotted] (39.69,-29.13) -- (38.77,-29.35) -- (39.48,-30.05);
		\draw [black,dotted] (31.4,-25.94) -- (39.7,-34.36);
		\fill [black,dotted] (39.7,-34.36) -- (39.49,-33.44) -- (38.78,-34.14);
		\draw [black,dotted] (32.12,-32.94) -- (38.98,-35.46);
		\fill [black,dotted] (38.98,-35.46) -- (38.41,-34.72) -- (38.06,-35.66);
		\draw [black,dotted] (32.21,-38.88) -- (38.89,-37.22);
		\fill [black,dotted] (38.89,-37.22) -- (37.99,-36.93) -- (38.23,-37.9);
		\draw [black,dotted] (44.7,-37.26) -- (50.8,-38.84);
		\fill [black,dotted] (50.8,-38.84) -- (50.15,-38.16) -- (49.9,-39.13);
		\draw [black,dotted] (44.6,-35.42) -- (50.9,-32.98);
		\fill [black,dotted] (50.9,-32.98) -- (49.98,-32.8) -- (50.34,-33.74);
		\draw [black,dotted] (43.85,-34.31) -- (51.65,-25.99);
		\fill [black,dotted] (51.65,-25.99) -- (50.74,-26.23) -- (51.47,-26.91);
		\draw [black,dotted] (44.7,-26.22) -- (50.8,-24.58);
		\fill [black,dotted] (50.8,-24.58) -- (49.9,-24.3) -- (50.16,-25.27);
		\draw [black,dotted] (44.57,-28.14) -- (50.93,-30.76);
		\fill [black,dotted] (50.93,-30.76) -- (50.38,-29.99) -- (50,-30.92);
		\draw [black,dotted] (43.86,-29.18) -- (51.64,-37.42);
		\fill [black,dotted] (51.64,-37.42) -- (51.45,-36.49) -- (50.73,-37.18);
		\draw [black,dotted] (56.16,-22.08) -- (62.94,-17.32);
		\fill [black,dotted] (62.94,-17.32) -- (62,-17.37) -- (62.58,-18.19);
		\draw [black,dotted] (56.7,-23.8) -- (62.4,-23.8);
		\fill [black,dotted] (62.4,-23.8) -- (61.6,-23.3) -- (61.6,-24.3);
		\draw [black,dotted] (56.17,-25.51) -- (62.93,-30.19);
		\fill [black,dotted] (62.93,-30.19) -- (62.56,-29.33) -- (61.99,-30.15);
		\draw [black,dotted] (56.21,-33.55) -- (62.89,-37.95);
		\fill [black,dotted] (62.89,-37.95) -- (62.5,-37.09) -- (61.95,-37.93);
		\draw [black,dotted] (55.49,-26.21) -- (63.61,-37.19);
		\fill [black,dotted] (63.61,-37.19) -- (63.54,-36.25) -- (62.74,-36.84);
		\draw [black,dotted] (55.02,-26.49) -- (64.08,-44.91);
		\fill [black,dotted] (64.08,-44.91) -- (64.17,-43.97) -- (63.27,-44.41);
		\draw [black,dotted] (55.45,-29.46) -- (63.65,-18.04);
		\fill [black,dotted] (63.65,-18.04) -- (62.78,-18.4) -- (63.59,-18.98);
		\draw [black,dotted] (56.17,-30.19) -- (62.93,-25.51);
		\fill [black,dotted] (62.93,-25.51) -- (61.99,-25.55) -- (62.56,-26.37);
		\draw [black,dotted] (56.7,-31.9) -- (62.4,-31.9);
		\fill [black,dotted] (62.4,-31.9) -- (61.6,-31.4) -- (61.6,-32.4);
		\draw [black,dotted] (55.49,-34.31) -- (63.61,-45.19);
		\fill [black,dotted] (63.61,-45.19) -- (63.53,-44.25) -- (62.73,-44.85);
		\draw [black,dotted] (55.01,-36.9) -- (64.09,-18.3);
		\fill [black,dotted] (64.09,-18.3) -- (63.29,-18.8) -- (64.18,-19.23);
		\draw [black,dotted] (55.49,-37.19) -- (63.61,-26.21);
		\fill [black,dotted] (63.61,-26.21) -- (62.74,-26.56) -- (63.54,-27.15);
		\draw [black,dotted] (56.21,-37.95) -- (62.89,-33.55);
		\fill [black,dotted] (62.89,-33.55) -- (61.95,-33.57) -- (62.5,-34.41);
		\draw [black,dotted] (56.7,-39.6) -- (62.4,-39.6);
		\fill [black,dotted] (62.4,-39.6) -- (61.6,-39.1) -- (61.6,-40.1);
		\draw [black,dotted] (56.18,-41.29) -- (62.92,-45.91);
		\fill [black,dotted] (62.92,-45.91) -- (62.55,-45.04) -- (61.98,-45.87);
		\end{tikzpicture}
		
	}
	\caption{Classic architecture for an AE. Black nodes denote a 2-variable encoding layer. Dark gray nodes are intermediate hidden layers. Light gray ones are the input (left) and output layers.}
	\label{Fig.AEStructure}
\end{figure}

\begin{figure*}[h!]\centering
  \includegraphics[width=.7\linewidth]{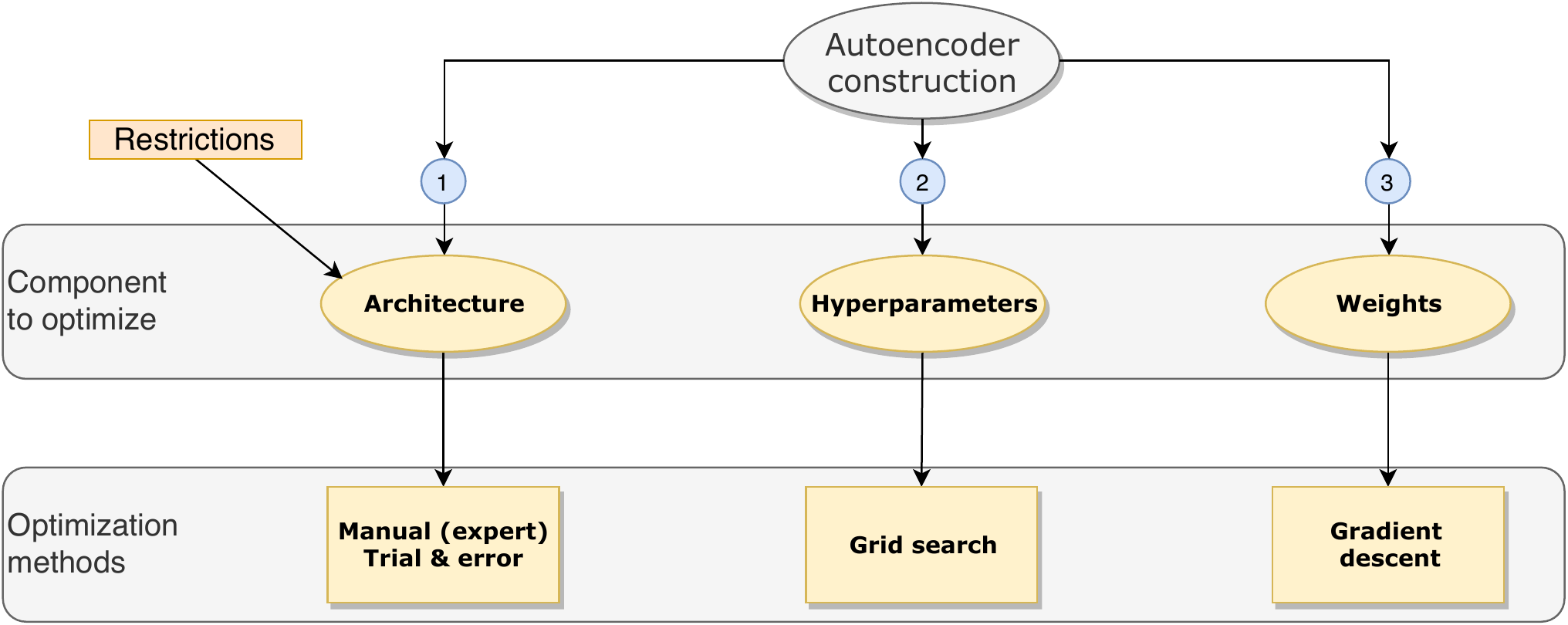}
  \caption{Components to be optimized during AE construction and the methods used for it. The numbers inside circles indicate the usual optimization order: firstly an architecture is set, then a set of hyperparameters is chosen, lastly the weights are adjusted.}
  \label{Fig.AEDesign}
\end{figure*}

Although AEs \hl{are mainly used} to perform feature fusion~\cite{Charte:ReviewAEs}, searching the manifold in which the parameters to rebuild the data are found, \hl{they have many other practical applications~\cite{Charte:ShowcaseAEs}. A properly configured AE can project data of any dimensionality into 2 or 3 dimensions so that patterns can be graphically visualized~\cite{yu2013embedding}. AEs can be used to detect anomalies~\cite{sakurada,park}, training them to faithfully reconstruct normal patterns. When anomalous data enters the AE, it produces a high output error denoting that these patterns do not follow the known distribution. Another interesting application of AEs, specifically of denoising AEs, is data noise removing. This kind of AE has been used to successfully denoise images~\cite{xie} and speech~\cite{speech}. Usually the loss function of the AE has to be adapted to the specific task to be faced, so that the obtained encoding promotes separability, topology preservation or any other desired characteristic. When only maximum performance is pursued while reconstructing the input patterns, the AE will produce a general purpose coding in its inner layer. } 

AEs can be configured with a variable amount of inner layers, each of them having different lengths. The proper architecture will mostly depend on the complexity of the patterns to be reconstructed and the restrictions imposed by the \hl{encoding} layer. These restrictions prevent the AE from simply copying the input onto the output~\cite{Charte:ReviewAEs}, for instance by reducing the number of neurons in the coding layer, forcing the output of most neurons to be zero (sparse representation), etc.

Finding the best parameters to tune a machine learning model is an uphill battle. Performing a grid search through an internal validation process is an usual approach. However, it is useful only for limited sets of parameters taking known ranges of values. \hl{Disparate search algorithms, metaheuristics~\cite{blum2003metaheuristics} and optimization approaches~\cite{adeli2006cost}, aimed to perform combinatorial optimization, could be applied. These go from relatively simple search methods~\cite{aarts1997local,den2001design,korf1990real,zhang1999algorithms}  such as A* or IDA to more advanced and complex approaches such as memetic algorithms~\cite{Neri2008}, including classic means as simulated annealing~\cite{van1987simulated} or tabu search~\cite{battiti1994reactive}.} Evolutionary methods~\cite{freitas2009review,Bck1993AnOO} have been also used to face optimization problems~\cite{wang2019optimizing,kyriklidis2016evolutionary,kociecki2014two,kociecki2015shape} for long time. \hl{Many of these algorithms are based on the behavior of certain populations, such as ant colonies~\cite{blum2002ant} and particle swarms~\cite{poli2007particle}. In the following, we focus specifically in approaches based on natural evolution~\cite{foster2001computational}.}

In evolutionary algorithms~\cite{back2018evolutionary} the search space is defined by the \textit{chromosome}. It is made up of several \textit{genes}, each of them representing a specific trait or dimension. The values taken by the genes in a chromosome are limited, and each combination is a potential solution (a point in the search space). These are also known as \textit{individuals}. Usually a set of these are randomly generated, producing an initial \textit{population}. Each individual is assigned a fitness value that represents the goodness of the solution. From this point, a number of iterations are run consisting in the same steps. Firstly, certain individuals of the population are selected for reproduction based on their fitness. Then, a set of operators are applied to selected individuals in order to create new ones. Common operators are crossing, that mixes genes of two individuals to produce a new chromosome, and mutation, which randomly changes the value of one or more genes. Lastly, the whole population is evaluated by computing the new fitness values and the population is updated, usually replacing old individuals with new ones, although the best overall solutions can be kept (elitism) whether they are new or not. \hl{Three} popular evolutionary methods are genetic algorithms~\cite{davis1991handbook}\hl{, differential evolution~\cite{storn1997differential} }and evolution strategy~\cite{rechenberg1984evolution}. \hl{The first follows the methodology just described of selection, crossing, mutation and evaluation. Although GAs can be seen as an old technique, they are still in widely use~\cite{adeli2006cost,kociecki2014two,kim2001discrete} due to their simplicity and good performance. The second method produces new individuals from differences between existing ones,} while the third one focuses on evolving only a few individuals using the mutation operator as only tool.

Evolutionary methods are also suitable for \hl{combinatorial optimization, and they have been used }for instance for support vector machines~\cite{friedrichs2005evolutionary} and more recently for deep learning networks~\cite{young2015optimizing}. Even though EMs have been already used to optimize ANNs, many of the proposals have been focused on learning the weights linked to each connection\hl{. This is known as conventional neuroevolution~\cite{ANNEvo1,ANNEvo2,ANNEvo3,ANNEvo4} and its main foundation is to use an EM instead of the traditional gradient descent algorithm to optimize weights. Therefore, a fixed network architecture is the base for all the population. The only difference among individuals is the set of weight matrices connecting each layer, values optimized by the EM. Neuroevolutive algorithms able to also evolve the network topology appeared later~\cite{floreano2008neuroevolution}, being aimed most of them to optimize MLPs. There are different approaches to construct the ANN architecture, being one of the most popular subnetworks composition~\cite{gruau1994automatic}. A recent survey in that matter can be found in~\cite{stanley2019designing}}.

\hl{More recently, }the fusion of different techniques to fully automate the process of choosing a proper model structure and hyperparameters, adjusting the weights, etc., has given birth to a new field known as AutoML~\cite{AutoML,ying2019bench,feurer2015efficient}. \hl{AutoML tools can be based on EMs, but also in Bayesian techniques and other optimization algorithms. Existing AutoML tools are mostly aimed to aid in the design of MLPs and CNNs~\cite{van2019automatic}, maybe the two most popular kind of ANNs. Essentially, these tools have a limited set of \textit{cells} or \textit{blocks} that they can combine to define the ANN topology. Depending on the task at glance a specific strategy is followed to choose these blocks. For instance, the AutoKeras tool~\cite{Jin2018AutoKerasAE} defines tasks that allow the automated design of ANNs for image, text and structured data classification. The ANN architecture is made up of predefined blocks such as ImageBlock, TextBlock or StructuredDataBlock, with some adjustable parameters. As consequence, these AutoML tools have a coarse granularity and only can be used to build some specific types of ANNs.} 

By contrast, in the present proposal EMs are used to evolve the architecture of AEs, \hl{a kind of ANN with some specific aspects such as its symmetric layout or its unsupervised learning approach. The AE layout is generated with a fine granularity, as described below, instead of using predefined blocks.} The weights are learned through the usual back-propagation algorithm.  The EvoAAA procedure proposed here could be a piece in an AutoML chain.

The rest of this paper is structured as follows: in Section \ref{Sec.AESearch} the different aspects to optimize while building an AE are outlined and the proposed architecture search methodology is presented. Section \ref{Sec.Experiments} describes the experimental framework, provides the analysis of performance results and discuss them. Lastly, conclusions are drawn in Section \ref{Sec.Conclusions}.

\section{Autoencoder architecture search with EvoAAA}\label{Sec.AESearch}

This section presents the proposal to face the problem formulated in subsection \ref{Sec.Problem}, outlining the aspects that have to be taken into account while building an AE and detailing the methodology to follow.

\subsection{Components to be optimized during AE construction}
The process to build a new AE, adjusted to produce a good representation from input patterns, implies several steps. Each one is linked to the optimization of a certain component. A summary of the procedure, components and methods is shown in Figure~\ref{Fig.AEDesign}. The components are tuned in the following order:

\begin{figure*}[h!]\centering
  \includegraphics[width=.9\linewidth]{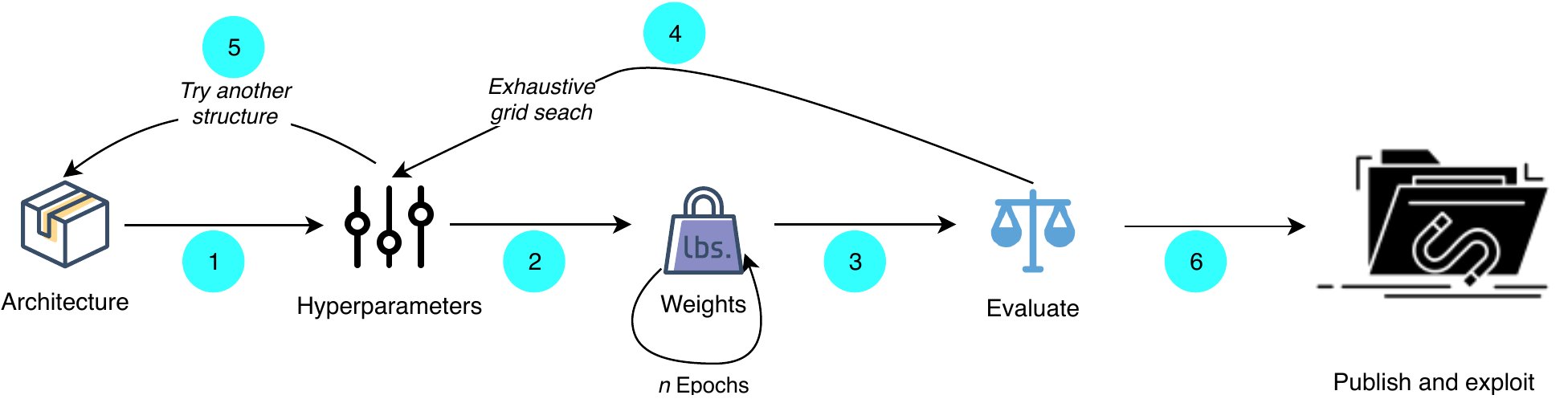}
  \caption{Before an AE can be exploited each architecture has to be tested with multiple hyperparameters combinations, and the weights of everyone of these configurations have to be adjusted.}
  \label{Fig.SearchProcess}
\end{figure*}

\begin{enumerate}
    \item \textbf{Architecture:} The structure of the AE has to be set, deciding the amount of layers it will be made \hl{of}, the number of units per layer, which activation functions will be used in each unit, etc. For years, this has been manually done by experts. The conventional trial and error procedure is also a frequent approach, readjusting the AE architecture after the three building steps have been completed and the AE performance evaluated. During this step specific restrictions, as the symmetrical structure of the network, have to be taken into account.
    
    \item \textbf{Hyper-parameters:} Having decided on the design of the AE, the following step would be choosing the proper values for several hyperparameters. These are not part of the network structure nor are they part of the weights configuration. They are in charge of controlling the tuning of weights during the training process. The most common hyperparameters are the learning rate, the batch size and the number of epochs the network is trained. Although the values could be manually picked, usually a grid search algorithm is used. Once more, trial and error using part of the training patterns as validation, allows to find the best configuration.
    
    \item \textbf{Weights:} Once the architecture of the network and its hyperparameters have been set, the last action would be adjusting the weights that connect every unit in each layer with all the units in the following one. Unlike what happens for finding a proper network structure and good hyperparameters, there is a solid mathematical background related to how these weights should be tuned. Derivatives allow to know the contribution of each connection to the global committed error, so that small adjustments can be made in the correct direction. This is the foundation of the well-known gradient descent method.
\end{enumerate}

The three previous steps are iterative in a nested way, as shown in Figure~\ref{Fig.SearchProcess}. Usually several different architectures will be tested. \hl{For each architecture, the first step (noted as 1 inside a circle) would be getting different sets of hyperparameters to be tried. Adjusting the weights of each configuration, a procedure that is iterative by itself, is be the following step. The third stage, after an architecture has been fixed, a set of hyperparameters is chosen and the weights are adjusted, is evaluating the model. The output of this action would lead us to step 4 or step 6, depending on whether a certain criterion is satisfied or not. In the first case additional sets of hyperparameters, generally obtained by grid search, would be used. If all potential combinations have been tried, a step backward (noted as 5) would be altering the AE architecture. Step 6 marks the end of the process, having an AE ready to be exploited in the system.} 

\subsection{The EvoAAA proposal}

As previously stated, there are well-known procedures for steps 2 and 3 (see Figure~\ref{Fig.AEDesign}) of the optimization process: grid search for finding the hyperparameters (step 2) and gradient descent for weight adjusting (step 3). On the contrary, finding a proper AE architecture is still an open problem. 

The experience acquired while working with a certain data set could not be applicable when data change occurs. The amount of combinations is potentially infinite, so automating this process by means of exhaustive search is unfeasible. Most experts follow some heuristics to choose the number of layers and units, depending on the quantity of variables they have to deal with, while aspects as activation functions are statically assigned.

Seeking a way of automating the AEs design process, a suitable architecture could be found through a simple search if the number of combinations is restricted beforehand. It would be similar to the grid search procedure followed for hyperparameters. This approach would allow to choose the best structure among a bag of predesigned ones. However, it does not offer any guarantee over the performance of these predesigned models. Other straightforward search heuristics could be adopted, such as adding/deleting layers/units from a base structure as long as the performance of the AE improves. 

The EvoAAA approach proposes to solve the task through EMs, since they have amply demonstrated their ability to solve disparate optimization problems. Specifically, it aims to use population based algorithms to evolve a set of AE architectures over time. To do so, a way to code all possible AE configurations is introduced. It will be the chromosome representation that the EMs will work with.

\begin{figure*}[htp]
  \includegraphics[width=\linewidth]{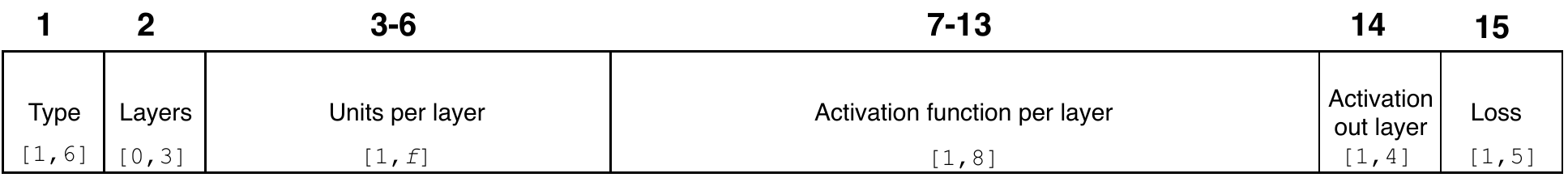}
  \caption{Chromosome genes, name and interval of values they can get.}
  \label{Fig.Chromosome}
\end{figure*}

\begin{table*}[t!]
\setlength{\tabcolsep}{6pt}
\renewcommand{\arraystretch}{1.5}
\caption{Purpose of each gene and description of their values.}\label{Table.Genes}
\scriptsize
\begin{tabular}{p{.10\textwidth}p{.35\textwidth}p{.45\textwidth}}
\toprule
\textbf{Name} & \textbf{Purpose} & \textbf{Values} \\
\midrule
Type & Sets the type of AE to be used & 1) Basic, 2) Denoising, 3) Contractive, 4) Robust, 5) Sparse, 6) Variational \\

Layers & Number of additional layers in coder/decoder & 0) Only a coding layer, 1-3) Additional layers in both coder and decoder\\

Units & Sets the number of units per layer, with $f$ being the amount of features in the dataset & The first integer (gen 3) configures the number of units in the outer layer, while the last one (gen 6) sets the coding length. \\

Activation & Activation function to use in each layer, both for the coder and decoder & 1) linear, 2) sigmoid, 3) tanh, 4) relu, 5) selu, 6) elu, 7) softplus, 8) softsign \\

Out act. & Activation function for the output layer & 1) linear, 2) relu, 3) elu, 4) softplus \\

Loss & Loss function to evaluate during fitting & 1) Mean squared error, 2) Mean absolute error, 3) Mean absolute percentage error, 4) Binary crossentropy, 5) Cosine proximity \\
\bottomrule
\end{tabular}
\end{table*}

\subsubsection{Chromosome representation}
Evolutionary algorithms usually work with binary or real-valued genes. A set of genes builds a chromosome or individual of the population. In EvoAAA each chromosome will code the complete architecture of an AE. However, an integer gene representation is used rather than binary or real-valued genes.

The chromosome will be made up of 15 genes, as shown in Figure~\ref{Fig.Chromosome}. The number of each gene is shown above, their names inside and just below the range of values that can be assigned to them. The purpose of each gene, as well as the meaning of its values, are portrayed in Table~\ref{Table.Genes}. The main characteristics of this AE \hl{encoding} are the following:

\begin{itemize}
    \item Different types of representations can be learned with AEs depending on the imposed restrictions. Some of those restrictions are linked to the type of AE~\cite{Charte:ReviewAEs}, designed to induce sparsity, learn from noisy samples, etc. The first gen in the chromosome allows to choose among six different AE types (see details in Table~\ref{Table.Genes}).
    
    \item The AEs will have a coding layer and up to six additional hidden layers that have to be taken in pairs: 2, 4 or 6. The value of the second gen, between 0 and 3, indicates the number of pairs of hidden layers. Therefore, the simplest AE would have only 3 layers, the input one, the coding one and output one, while the most complex would be made up of 9 layers in total.
    
    \item Genes 3 to 6 state the number of units to have in the hidden layers. The last value is associated to the innermost layer, so it sets the coding length. The other three values are linked to each layer pair, from outer to inner. The number of features in the data, noted as $f$, will limit the amount of units in any layer. An additional restriction is that inner layers cannot be larger than outer ones.
    
    \item With the exception of the input layer, which is limited to transferring values to the next one, all other layers in the AE use an activation function. Although it is common for all units of an AE to have the same activation function, there is nothing that restricts the use of different functions. The proposed \hl{encoding} uses 7 genes (7 to 13) to choose the activation functions to be applied in each inner layer, allowing eight different options (see Table~\ref{Table.Genes}). The output layer is treated independently with gene 14, since only activation functions producing positive output are allowed.
    
    \item The last gene in the chromosome can take values from 1 to 5, stating the loss function to be internally evaluated while training the AE. The meaning of these values is provided in Table~\ref{Table.Genes}.
\end{itemize}

Through the restrictions imposed in genes 3 to 6, the resulting AEs would be always undercomplete~\cite{Charte:ReviewAEs}. This means that the learned representation will be more compact than the original one, with a vector containing less values. In addition, the loss function evaluated while adjusting the weights will exclusively focus on reconstruction error. AEs can be trained to improve class separability, reduce the data complexity and other goals. In this case the main objective is to have a reconstruction of data patterns as good as possible. This will force the inner layer to concentrate as much general purpose useful information as possible, so that the AE can be later used in any kind of task.

\subsubsection{Autoencoder complexity penalization}\label{Sec.Penalization}

When designing an AE for learning new representations the main interest will be in the raw performance. In an AE the performance is usually measured as the error produced while reconstructing data patterns from the learned encoding. Nonetheless, the time needed to obtain the encoding is also a factor to consider. This is the motivation to include a penalization factor in EvoAAA: $\alpha$. It will be used in the fitness function of the evolutionary method in order to assess the goodness of each AE configuration.

The fitness function, that will decide the quality of the solutions, will be computed as shown in (\ref{Eq.fitness}), where $trainloss$ is the reconstruction loss produced by the AE with training data, $Layers$ is the number of additional hidden layers (gen 2), $Units~coding$ is the size of the \hl{encoding} layer (gen 6), and $\alpha$ is a coefficient setting the level of penalization applied according to the complexity of the AE. 

\begin{equation}
fitness = train loss + \alpha(Layers\times Units~coding)
\label{Eq.fitness}
\end{equation}

Therefore, AEs having a similar reconstruction performance but simpler architecture \hl{(see subsection~\ref{Sec.PenalizationAnalysis} for additional details)} will be preferred over the more complex ones. The bias to obtain AEs with less layers and a shorter encoding is modulated with the $\alpha$ value. This is the only parameter needed by EvoAAA.

\subsubsection{Search space}

The reason for using a complex evolutionary algorithm to find a proper AE architecture is that the search space is huge. So, a strategy is needed to pick a good solution without having to explore much of this space. But, how large is the search space assuming the chromosome previously described?

Excluding genes 3-6, whose values would vary depending on the number of features in the dataset, there are more than a billion combinations (see Eq.~\ref{Eq.combinations}). 

\begin{equation}
\begin{array}{l}
Type\times Layers\times Act^7\times Act~out\times Loss =\\
\\
6\times 4\times 8^7\times 4\times 5 = 1~006~632~960
\end{array}
\label{Eq.combinations}
\end{equation}

\hl{Working with very small }datasets\hl{, those} having a few dozens of attributes only, \hl{the amount of combinations }will grow to several billions. \hl{We would face} trillions of solutions or even more for high-dimensional datasets. Evaluating all those solutions to find the best one is currently unfeasible. \hl{As a result, looking for an optimal AE architecture would not }be always possible by brute force. However, we could find good enough solutions through optimization mechanisms based on evolutionary strategies.

\section{Algorithmic framework of EvoAAA}\label{Sec.Experiments}

The EvoAAA methodology is a general evolutionary proposal to search good AE architectures. It can be instantiated using any population based evolutionary algorithm. In this study \hl{three} specific instances of EvoAAA are tested, one with a genetic algorithm as underlying search method, \hl{another one relying on differential evolution} and the \hl{third one} using evolution strategy. \hl{The three of them} will be compared against each other and also versus \hl{two baseline search methods, a} simple exhaustive search \hl{and a random search}. Nine different data sets will be used in the conducted experimentation.

\subsection{Evolutionary search algorithms}

We propose instantiating EvoAAA \hl{three times} using \hl{three} different evolutionary algorithms. \hl{All} of them will use the former chromosome representation. These \hl{three} approaches are:

\begin{itemize}
	\item \textbf{Genetic algorithm (GA).} A classical genetic algorithm~\cite{davis1991handbook}, in which a population of individuals evolves through a crossover operator, to give rise to new ones, and to which a mutation operator is applied with a certain probability. 
	
	\item \textbf{Evolution strategy (ES).} An aggressive solution-seeking algorithm~\cite{rechenberg1984evolution}, working with a few individuals who give rise to new ones exclusively through mutation.
	
	\hl{\item \textbf{Differential evolution (DE).} A population-based optimization algorithm~\cite{storn1997differential} in which new individuals (\textit{agents}) are produced from the differences between two random agents with respect to another taken as reference.}
\end{itemize}

\begin{table}[t]
	\caption{\hl{Main parameters of the evolutionary algorithms.}}\label{Table.Parameters}
	\setlength{\tabcolsep}{12pt}
	\centering\footnotesize
	\begin{tabular}{lrrr}
		\toprule
		\textbf{Parameter} & \textbf{GA} & \textbf{ES} & \textbf{DE} \\
		\midrule
		Population size 		&  50  &   4   &  150 \\
		Iterations      		& 100  & 500   &   30 \\
		Prob. mutation  		& 1/15 & 1/15  &   NA \\
		Prob. cross-over        & 1.0    & NA    &  0.5 \\
		Elitism (individuals)  	&   5  & NA    &   NA \\
		Termination cost 		&   0  & 0     &    0 \\
		\bottomrule
	\end{tabular}
\end{table}

Table~\ref{Table.Parameters} summarizes the main parameters used to run these methods.  \hl{For GA and ES, e}ach gene in the chromosome is mutated with a probability of 1/15, a value based on the chromosome length itself. Elitism is used in the GA to preserve the tenth percent of individuals having better fitness. The ES is intrinsically elitist, choosing only the best candidates among the union of new mutated individuals and the old population. For GA, in each iteration the best 5 individuals are chosen and preserved. Then, two random parents are picked up from the remaining individuals by using roulette wheel probability. A multi-point crossover operator is applied, being the crossing points established according to the diagram in Figure~\ref{Fig.Chromosome}. This allows the two individuals acting as parents to interchange several of their genes to produce childhood. This way, new individuals are produced until the population size is met, replacing all the old individuals. Lastly, the mutation operator is used with some individuals according to the probability previously stated. As can be seen, this is an aggressive scheme that maximizes the exploration in such a huge search space. \hl{For DE, the DE/local-to-best/1 optimization approach is followed. The population size is obtained from the representation length (15 genes $\times$ 10). Cross-over probability, as well as other specific parameters, has been set following the recommendations in \cite{storn1997differential,price2006differential}. The number of iterations for each method has been adjusted so that a similar amount of evaluations is made}.

\subsection{Data sets}

\begin{table}[t]
	\centering
	\caption{Datasets used in the experimental study} 
	\begin{tabular}{llrrc}
		\toprule
		\textbf{Dataset} & \textbf{Type} & \textbf{Variables} & \textbf{Instances} & \textbf{Source} \\ 
		\midrule
		cifar10 & Integer & 1~024 & 60~000 & \cite{cifar10}\\ 
		delicious & Binary & 983 & 16~105 & \cite{delicious}\\ 
		fashion & Integer & 784 & 70~000 & \cite{fashion} \\ 
		glass &  Real &  9 & 214 & \cite{glass} \\ 
		ionosphere & Real & 34 & 351 & \cite{ionosphere} \\ 
		mnist & Integer & 784 & 70~000 & \cite{mnist} \\ 
		semeion & Binary & 256 & 1~593 & \cite{semeion}\\ 
		sonar & Real & 60 & 208 & \cite{sonar}\\ 
		spect & Binary & 22 & 267 & \cite{spect}\\ 
		\bottomrule
	\end{tabular}
    \label{Table.Datasets}
\end{table}

In order to compare the performance of the \hl{three} instances of EvoAAA \hl{along with the exhaustive and random search approaches}, nine data sets are used. Their main traits are detailed in Table~\ref{Table.Datasets}. The last column provides the origin reference for each one of them. The criteria followed to choose them have been:

\begin{itemize}
	\item \textbf{Attribute type:} An AE is a specialized ANN that, through a series of computations, reconstructs the input values. These have to be numeric, and three cases are considered: real values, integer values, and binary values. The goal is to evaluate how the AEs performance changes depending on the type of attributes they have to reconstruct.
	
	\item \textbf{Number of attributes:} Half of the data sets have several hundreds of attributes, even more than a thousand in the case of cifar10. The other five are not that large, with only a handful of variables. This way the behavior of found AEs while working with a few versus a lot of variables will be analyzed.
\end{itemize}

Since an AE is an unsupervised representation learning method, class attributes have been removed for all the data sets. The number of variables indicated in Table~\ref{Table.Datasets} is the effective amount of them being processed.

\subsection{Restrictions and evaluation}

\hl{Five} AE architectures will be obtained for each data set, EvoAAA-Gen: the instance based on a genetic algorithm, EvoAAA-Evo: the instance using evolutionary strategy as underlying search method\hl{, EvoAAA-Dif: the one based on differential evolution, and the two produced by the exhaustive (Exh) as well as random search (Ran)}. The same restrictions and evaluation procedure are applicable to all of them:

\begin{itemize}
	\item \textbf{Performance evaluation:} The common mean squared error metric is used to assess the performance of the AEs. This is an unbounded measure, which depends on the original range of values. The lower the MSE the better performance the AE has. This value is the $trainloss$ factor in Eq.~\ref{Eq.fitness}.

	\item \textbf{Termination cost:} As shown in Table~\ref{Table.Parameters}, the \hl{three} evolutionary algorithms have been configured with 0 as termination cost. This means that the search will stop if a configuration returning $MSE=0$ is found, but not before unless the maximum runtime \hl{or maximum number of iterations are} reached.
	
	\item \textbf{Maximum runtime:} The \hl{five} search approaches will be limited to running for 24 hours. After that the best solution found until then is returned as preferred AE structure. \hl{In practice, this limitation will impede the exhaustive search from going through all the possible AE configurations, limiting as well the amount of combinations tested by the random search.}
\end{itemize}

Aside from the MSE, the amount of architectures tested by each approach is also recorded during execution, as well as each individual solution. The number of combinations is limited by both the maximum runtime and the population size and iterations of the evolutionary methods.

\subsection{Experiments and results}

\begin{table*}[t!]
\centering
\caption{\hl{Summary of results. For each combination dataset-optimization strategy the amount of evaluated individuals, number of layers and encoding length of the AE, its complexity and MSE (including the penalization factor) are provided.}}
\label{Table.Results}
\renewcommand{\arraystretch}{1.1}
\begin{tabular}{lrrrrrr}
  \toprule
\textbf{Dataset} & \textbf{Method} & \textbf{Individuals} & \textbf{Layers} & \textbf{Coding length} & \textbf{Complexity} & \textbf{Error (MSE)} $\downarrow$ \\ 
  \midrule
  cifar10 
  & EvoAAA-Dif & 2~537 & 1 & 38 & 0.004 & 0.0137 \\
  & EvoAAA-Evo & 4~001 & 1 & 52 & 0.005 & 0.0133 \\ 
  & Exh & 1~739 & 1 &  1 & 0.000 & 0.0395 \\ 
  & EvoAAA-Gen & 1~115 & 1 & 39 & 0.004 & \textbf{0.0131} \\ 
  & Ran & 2~101 & 3 &  5 & 0.002 & 0.0260 \\
\cmidrule{1-2}
  delicious 
  & EvoAAA-Dif & 4~650 & 2 & 19 & 0.004 & 0.0124 \\
  & EvoAAA-Evo & 4~001 & 1 & 39 & 0.004 & 0.0119 \\ 
  & Exh & 5~837 & 1 & 1 & 0.000 & 0.0165 \\ 
  & EvoAAA-Gen & 1~401 & 1 & 32 & 0.003 & \textbf{0.0102} \\ 
  & Ran & 5~901 & 2 & 10 & 0.002 & 0.0134 \\
\cmidrule{1-2}
  fashion 
  & EvoAAA-Dif & 2~604 & 1 & 256 & 0.026 & \textbf{444.8169} \\
  & EvoAAA-Evo & 755 & 3 & 61 & 0.012 & 565.4571 \\ 
  & Exh & 1~494 & 1 & 1 & 0.000 & 4~180.2370 \\ 
  & EvoAAA-Gen & 439 & 5 & 156 & 0.047 & 1~881.0680 \\ 
  & Ran & 1~501 & 2 & 35 & 0.007 & 782.3572 \\
\cmidrule{1-2}
  glass 
  & EvoAAA-Dif & 4~650 & 2 & 3 & 0.001 & 24.8785 \\
  & EvoAAA-Evo & 4~001 & 3 & 3 & 0.001 & 5.4093 \\ 
  & Exh & 35~368 & 1 & 1 & 0.000 & 29.2589 \\ 
  & EvoAAA-Gen & 4~505 & 1 & 5 & 0.000 & \textbf{0.4473} \\ 
  & Ran & 35~301 & 3 & 7 & 0.002 & 1.2705 \\
\cmidrule{1-2}
  ionosphere 
  & EvoAAA-Dif & 4~650 & 2 & 23 & 0.005 & \textbf{0.0740} \\
  & EvoAAA-Evo & 4~001 & 3 & 15 & 0.003 & 0.0931 \\ 
  & Exh & 29~905 & 1 & 1 & 0.000 & 0.2099 \\ 
  & EvoAAA-Gen & 2~302 & 3 & 15 & 0.003 & 0.0917 \\ 
  & Ran & 29~901 & 3 & 23 & 0.007 & 0.1049 \\
\cmidrule{1-2}
  mnist 
  & EvoAAA-Dif & 2~754 & 1 & 162 & 0.016 & \textbf{192.5133} \\
  & EvoAAA-Evo & 732 & 3 & 332 & 0.066 & 431.6960 \\ 
  & Exh & 1~491 & 1 & 1 & 0.000 & 4~104.1400 \\ 
  & EvoAAA-Gen & 403 & 1 & 156 & 0.016 & 254.7397 \\ 
  & Ran & 1~401 & 2 & 35 & 0.007 & 611.5036 \\
\cmidrule{1-2}
  semeion 
  & EvoAAA-Dif & 4~650 & 1 & 136 & 0.014 & 0.0459 \\
  & EvoAAA-Evo & 4~001 & 1 & 134 & 0.013 & \textbf{0.0355} \\ 
  & Exh & 19~861 & 1 & 1 & 0.000 & 0.1940 \\ 
  & EvoAAA-Gen & 4~505 & 1 & 110 & 0.011 & 0.0376 \\
  & Ran & 19~901 & 2 & 132 & 0.026 & 0.0604 \\ 
\cmidrule{1-2}
  sonar 
  & EvoAAA-Dif & 4~650 & 1 & 29 & 0.003 & 0.0142 \\
  & EvoAAA-Evo & 4~001 & 3 & 17 & 0.003 & 0.0162 \\ 
  & Exh & 34~557 & 1 & 1 & 0.000 & 0.0462 \\ 
  & EvoAAA-Gen & 4~505 & 3 & 7 & 0.001 & \textbf{0.0139} \\ 
  & Ran & 35301 & 4 & 8 & 0.003 & 0.0145 \\
\cmidrule{1-2}
  spect 
  & EvoAAA-Dif & 4~650 & 2 & 13 & 0.003 & 0.0959 \\
  & EvoAAA-Evo & 4~001 & 3 & 14 & 0.003 & 0.0834 \\ 
  & Exh & 30~740 & 1 & 1 & 0.000 & 0.1829 \\ 
  & EvoAAA-Gen & 4~505 & 5 & 15 & 0.004 & \textbf{0.0703} \\ 
  & Ran & 30~701 & 3 & 14 & 0.004 & 0.1145 \\
  \bottomrule
\end{tabular}
\end{table*}

The \hl{45} experiments\footnote{\hl{The R code to reproduce these experiments, as well as the datasets used in them, are available to download from \url{https://github.com/fcharte/EvoAAA}. To execute this code you will need a current R version, install several R packages (detailed in the provided scripts) and the frameworks enumerated in this section. Each run will provide two result files containing all evaluated solutions and the AE structure for the best ones.}} (9 data sets times \hl{5} search approaches) have been executed in the same hardware, a PC with an NVidia RTX-2080 GPU, and using the following software configuration: Arch Linux~\cite{ArchLinux}, CUDA 10.1~\cite{cuda}, cudnn 7.6~\cite{cudnn}, Tensorflow 1.14~\cite{tensorflow}, Keras 2.24~\cite{keras} and ruta 1.1~\cite{Charte:rutaPackage}. The RMSProp~\cite{RMSProp} optimizer of Tensorflow has been used. Since it relies on an adaptive learning rate, optimizing this hyper-parameter is not necessary. The default batch size value in Keras has been chosen, 32. Lastly, the training of the AEs is made in 20 epochs. Although a higher number of epochs, such as 100 or even more, is usual while trying to perform a final optimization of a representation, in this case the main interest is in comparing the search procedures rather than in obtaining fully optimized AEs. Reducing the number of epochs allows to try more architectures in a certain time interval. Regarding how the data instances are used, 20\% of them are reserved from the beginning to assess AE performance, computing the MSE. The remainder 80\% patterns are given to Keras to train each AE architecture.

A summary of results is provided in Table~\ref{Table.Results}. For each data set the columns indicate its name, the search method, the amount of individuals (AE configurations) tested, the number of hidden layers and encoding length of the best individual, its complexity (assuming that $\alpha=0.0001$) and the MSE obtained with test data. For these last two columns the lower the value the better the AE would be, having less complexity and superior reconstruction performance. The best configuration (lower MSE) for each data set has been highlighted in bold. These values include the complexity penalization, real best MSEs are provided in Table~\ref{Table.Averages}. \hl{Average error values, computed from all the evaluated solutions, and best values are presented in it.}

The number of tested individuals is limited by the population size and iterations \hl{, as well as the maximum runtime,} for the \hl{three} EvoAAA instances. In addition all the invalid configurations produced during the exploration are discarded before they enter the training and testing phase. The exhaustive \hl{and random strategies} only generate valid AE architectures, and the only limitation is the running time.  

\begin{table*}[t!]
\centering
\caption{\hl{Average and best performance (raw MSE) by data set and search strategy}}
\label{Table.Averages}\footnotesize
\begin{tabular}{lrrrrrrrrr}
  \toprule
 & \textbf{cifar10} & \textbf{delicious} & \textbf{fashion} & \textbf{glass} & \textbf{ionosphere} & \textbf{mnist} & \textbf{semeion} & \textbf{sonar} & \textbf{spect} \\ 
  \midrule
  EvoAAA-Dif (Avg.) & 0.0256 & 0.0186 & 8979.9337  & 604.3130 & 0.2719 & 5073.9444 & 0.2348 & 0.0404 & 0.2575 \\ 
  EvoAAA-Evo (Avg.) & \textbf{0.0085} & 0.0086 & \textbf{891.1958}   & 562.3818 & \textbf{0.1152} &  \textbf{616.6640} & \textbf{0.0322} & 0.0223 & 0.1145 \\ 
  Exh (Avg.) & 0.0498 & 0.0195 & 18734.2379 & 611.4017 & 0.3687 & 9646.3325 & 0.2836 & 0.1331 & 0.3213 \\ 
  EvoAAA-Gen (Avg.) & 0.0097 & \textbf{0.0075} & 2852.7007  & \textbf{345.4758} & 0.1227 & 3323.4105 & 0.0405 & \textbf{0.0157} & \textbf{0.1006} \\ 
  Rnd (Avg.) & 0.0461 & 0.0206 & 8557.0936  & 606.7591 & 0.3303 & 4997.0538 & 0.2886 & 0.0761 & 0.2910 \\ 
  \cmidrule{1-1}
  EvoAAA-Dif (Best) & \textbf{0.0020} & 0.0131 & \textbf{444.7913}   & 563.1547 & \textbf{0.0694} &  \textbf{192.4971} & 0.0323 & \textbf{0.0106} & 0.0924 \\ 
  EvoAAA-Evo (Best) & 0.0080 & 0.0080 & 565.4449   & 316.0380 & 0.0963 &  431.6296 & \textbf{0.0221} & 0.0128 & 0.0806 \\ 
  Exh (Best) & 0.0394 & 0.0164 & 4180.2370  & 576.7005 & 0.2098 & 4104.1400 & 0.1939 & 0.0461 & 0.1828 \\ 
  EvoAAA-Gen (Best) & 0.0090 & 0.0055 & 1881.0210  &   \textbf{0.4468} & 0.0872 &  254.7241 & 0.0266 & 0.0115 & \textbf{0.0658} \\ 
  Rnd (Best) & 0.0047 & \textbf{0.0021} & 782.3502   & 564.4037 & 0.0980 &  564.2114 & 0.0339 & 0.0114 & 0.1098 \\ 
   \bottomrule
\end{tabular}
\end{table*}

\begin{table*}[t!]
\centering
\caption{Performance ranking of the tested methods}
\label{Table.Ranking}
\hl{
\begin{tabular}{lrrrrr}
  \toprule
\textbf{Dataset} & \textbf{EvoAAA-Dif} & \textbf{EvoAAA-Evo} & \textbf{Exh} & \textbf{EvoAAA-Gen} & \textbf{Ran} \\ 
  \midrule
cifar10 & 1 & 3 & 5 & 4 & 2 \\ 
  delicious & 4 & 3 & 5 & 2 & 1 \\ 
  fashion & 1 & 2 & 5 & 4 & 3 \\ 
  glass & 3 & 2 & 5 & 1 & 4 \\ 
  ionosphere & 1 & 3 & 5 & 2 & 4 \\ 
  mnist & 1 & 3 & 5 & 2 & 4 \\ 
  semeion & 3 & 1 & 5 & 2 & 4 \\ 
  sonar & 1 & 4 & 5 & 3 & 2 \\ 
  spect & 3 & 2 & 5 & 1 & 4 \\ 
  \cmidrule{1-1}
  \textbf{Average} & 2.00 & 2.56 & 5.00 & 2.33 & 3.11 \\
   \bottomrule
\end{tabular}
}
\end{table*}

\subsection{Discussion}

The results shown in Table~\ref{Table.Results} can be analyzed from several perspectives, raw performance: the lower MSE the better; size of explored space: although the more inspected individuals could be considered the better, the ability to find good solutions exploring less space has to be also taken into account, and solution complexity: the fewer layers and shorter encoding length the better. In addition, other aspects such as running times, convergence, etc., can be studied.

\subsubsection{Analysis of performance}

It is easy to see that the highest error values always correspond to the exhaustive search approach. MSE is unbounded and depends on the original attribute values. So, it does not allow to make comparisons among different data, but only between the \hl{five} methods for each data set. In general, the error committed by \textbf{Exh} is one or two orders of magnitude above \hl{{EvoAAA-Dif},} {EvoAAA-Gen} and {EvoAAA-Evo}.

Comparing the \hl{three} EvoAAA instances, {EvoAAA-Gen} is the best performer in \hl{5} out of 9 datasets, \hl{with {EvoAAA-Dif} in a close second position gaining 3 out of 9 cases} although the differences against {EvoAAA-Evo} are almost negligible in some cases. The biggest difference can be found with the fashion dataset, where the \hl{EvoAAA-Dif} approach achieves one-\hl{quarter} of the error shown by EvoAAA-Gen while using a simpler architecture. \hl{This analysis is made taking into account the complexity of AEs, i.e. the MSE is increased by the penalization factor.}

In addition to best values, it would also be interesting to analyze the average behavior of the search strategies. For doing so, Table~\ref{Table.Averages} shows for each data set (columns) and method (rows) the average and minimum (best) MSE\footnote{These are raw MSE values rather than MSE penalized by complexity (the penalization factor was explained in subsection~\ref{Sec.Penalization} and it depends on the $\alpha$ parameter, set to $0.0001$ in the described experiments). Therefore, best values in Table~\ref{Table.Averages} are lower than those in Table~\ref{Table.Results}.} achieved in each case. It can be observed that the best MSE obtained by the exhaustive \hl{and random} search is far from the average MSE of \hl{{EvoAAA-Dif}, } {EvoAAA-Gen} and {EvoAAA-Evo}. This leads to the conclusion that even a few iterations with EvoAAA would achieve a better result than 24h of exhaustive \hl{or random} search. Internal differences between best and average values for each search method tend to be minimal in most cases (cifar10, delicious, ionosphere, semeion and sonar). In general, these differences seem lower in the case of {EvoAAA-Evo} than with {EvoAAA-Gen} \hl{or {EvoAAA-Dif}}. For instance, average and best values for fashion, glass and mnist are closer in the former case than in the latter. This suggests that {EvoAAA-Evo} would be preferable if only a few iterations are affordable.

\hl{If we strictly focus on the best values achieved by each approach, these returned at the end of all iterations without complexity penalization, {EvoAAA-Dif} stands out over the other algorithms. It has 5 best values, against 2 for {EvoAAA-Gen}, 1 for {EvoAAA-Evo}, and 1 for the random search. To assess the overall performance of each method a ranking is provided in Table~\ref{Table.Ranking}, with the last row showing the average rank. As can be seen,  {EvoAAA-Dif} is the best performer, closely followed by {EvoAAA-Gen} and  {EvoAAA-Evo}. The rank difference between random search and exhaustive search is considerable. The bias in the exhaustive search, which tries consecutive configurations until the runtime is out, implies less opportunities to explore the solution space than the random search. By applying a Friedman statistical test over the best MSE values (bottom half of Table~\ref{Table.Averages}) a $\textit{p-value}=0.0004248178$ was obtained. This means that there are statistically significant differences among the evaluated optimization strategies.}

\subsubsection{Analysis of explored space}

As can be seen in Table~\ref{Table.Results}, in general the exhaustive \hl{and random }search \hl{approaches} explore a much larger space of solutions than the evolutionary methods. In some cases, such as glass, ionosphere, semeion, sonar and spect, this approach examines up to 15 times more configurations than EvoAAA. This is  due to that \hl{these search algorithms} devote all their time in analyzing candidate solutions while the evolutionary methods \hl{have} other tasks to do, such as individuals selection, crossing, mutation, etc. However, \hl{neither random nor }exhaustive search never achieve the best performance \hl{when AE complexity is taken into account}. On the contrary, the MSE for \hl{these methods} is always higher. \hl{Exhaustive search} gets stuck in simpler architectures, as stated by its complexity values, despite its usually longer run times (analyzed later) that always go to 24 hours. This behavior is due to the way the exhaustive search has been implemented, trying all possible solutions allowed by the representation in Figure~\ref{Fig.Chromosome} from right to left as is usually done when an interval of values is going to be traversed from beginning to end. \hl{Random search has the ability to explore all the solution space, but it is a non-guided approach. In a way the strategies followed by the exhaustive and random search are antagonistic. The former chooses to exploit the local space, trying all the possible solutions of a very reduced area. The second one jumps all over the space of solutions, without taking advantage of past configurations to exploit the locality of the most promising solutions.}

The amount of solutions explored by the \hl{{EvoAAA-Dif}, } {EvoAAA-Gen} and {EvoAAA-Evo} methods is quite similar while working with small datasets, such as glass, semeion, sonar or spect. By contrast, the {EvoAAA-Evo} \hl{and {EvoAAA-Dif} approaches are} able to inspect more candidate structures when larger datasets are used. This \hl{could be} due to the simpler procedure of {EvoAAA-Evo} to produce its offspring with respect to {EvoAAA-Gen}, since crossing is not necessary and the population size is smaller, \hl{and to the lower number of iterations conducted by {EvoAAA-Dif} with respect to {EvoAAA-Gen}.}

\subsubsection{Analysis of solution complexity}

Another fact to take into account while comparing the different search procedures is the complexity of found architectures. Theoretically, simpler architectures achieving a similar performance would be preferable to more complex ones. 

As can be stated by looking at the sixth column in Table~\ref{Table.Results}, the lowest complexity is always that of the \textbf{Exh} approach. As said before, the exhaustive search is stuck in small architectures with thousands of combinations for activation functions and loss functions (see Figure~\ref{Fig.Chromosome}). However, these are not the best solutions as the MSE values demonstrate.

\hl{As would be expected, the random search also produces disparate AE configurations. Sometimes are simpler than those produced by the EvoAAA methods and sometimes are more complex.}

Despite some exception, such as fashion and spect, {EvoAAA-Gen} usually produces simpler AEs with fewer layers and a more compact encoding than \hl{{EvoAAA-Dif} and} {EvoAAA-Evo}. Therefore, at first sight the {EvoAAA-Gen} search seems to be the best choice, as it provides simpler AEs with better reconstruction capability. However, this comes with a cost as explained below.

\subsubsection{Analysis of running times}

The running times recorded during the experiments for each configuration are provided in Table~\ref{Table.Runtimes}. As in Table~\ref{Table.Results}, for each data set there are \hl{five} rows corresponding to the \hl{five} search approaches. Analyzing this information the following conclusions can be drawn:

\begin{table*}[h!]
	\centering
	\caption{\hl{Running time summary}}
	\label{Table.Runtimes}
	\begin{tabular}{lccccc}
		\toprule
		& \multicolumn{5}{c}{\textbf{Running times} $\downarrow$}  \\
		\cmidrule{2-6}
		\textbf{Dataset} & \textbf{EvoAAA-Dif} & \textbf{EvoAAA-Evo} & \textbf{Exh} & \textbf{EvoAAA-Gen} & \textbf{Rnd} \\
		\midrule
		cifar10 &  24h &  24h &  24h &  24h &  24h \\ 
		delicious & 18h 4m. & 17h 54m. &  24h & 6h 51m. &  24h \\ 
		fashion &  24h &  24h &  24h &  24h &  24h \\ 
		glass & 1h 41m. & 1h 20m. &  24h & 2h 20m. &  24h \\ 
		ionosphere & 1h 33m. & 0h 44m. &  24h & 1h 59m. &  24h \\ 
		mnist &  24h &  24h &  24h &  24h &  24h \\ 
		semeion & 2h 28m. & 1h 32m. &  24h & 5h 0m. &  24h \\ 
		sonar & 1h 24m. & 1h 10m. &  24h & 4h 13m. &  24h \\ 
		spect & 1h 31m. & 0h 59m. &  24h & 5h 57m. &  24h \\ 
		\bottomrule
	\end{tabular}	
\end{table*}

\begin{itemize}
    \item The exhaustive \hl{and random strategies} always reach the limit of 24 hours without being able to examine enough configurations to find a good solution\hl{, although the random approach is close to the evolutionary methods in some cases.}
    
    \item For the larger data sets (cifar10, fashion and mnist) none of the \hl{five} approaches finish the search process, running out of time.
    
    \item In general, \hl{{EvoAAA-Dif} and }{EvoAAA-Evo} only need a fraction of the time used by {EvoAAA-Gen} to find AE architectures slightly more complex but with a similar performance.
\end{itemize}

From this analysis a simple guideline can be drawn, choose the  \hl{{EvoAAA-Dif}} EvoAAA instance if performance is the main and only goal, but consider the {EvoAAA-Evo} instance if running time is important while sacrificing a bit of reconstruction power. For large data sets the latter approach can save many hours in obtaining a good-enough AE architecture.

\subsubsection{Solutions explored over time}

\begin{figure}[ht!]\centering
	\includegraphics[width=.5\linewidth]{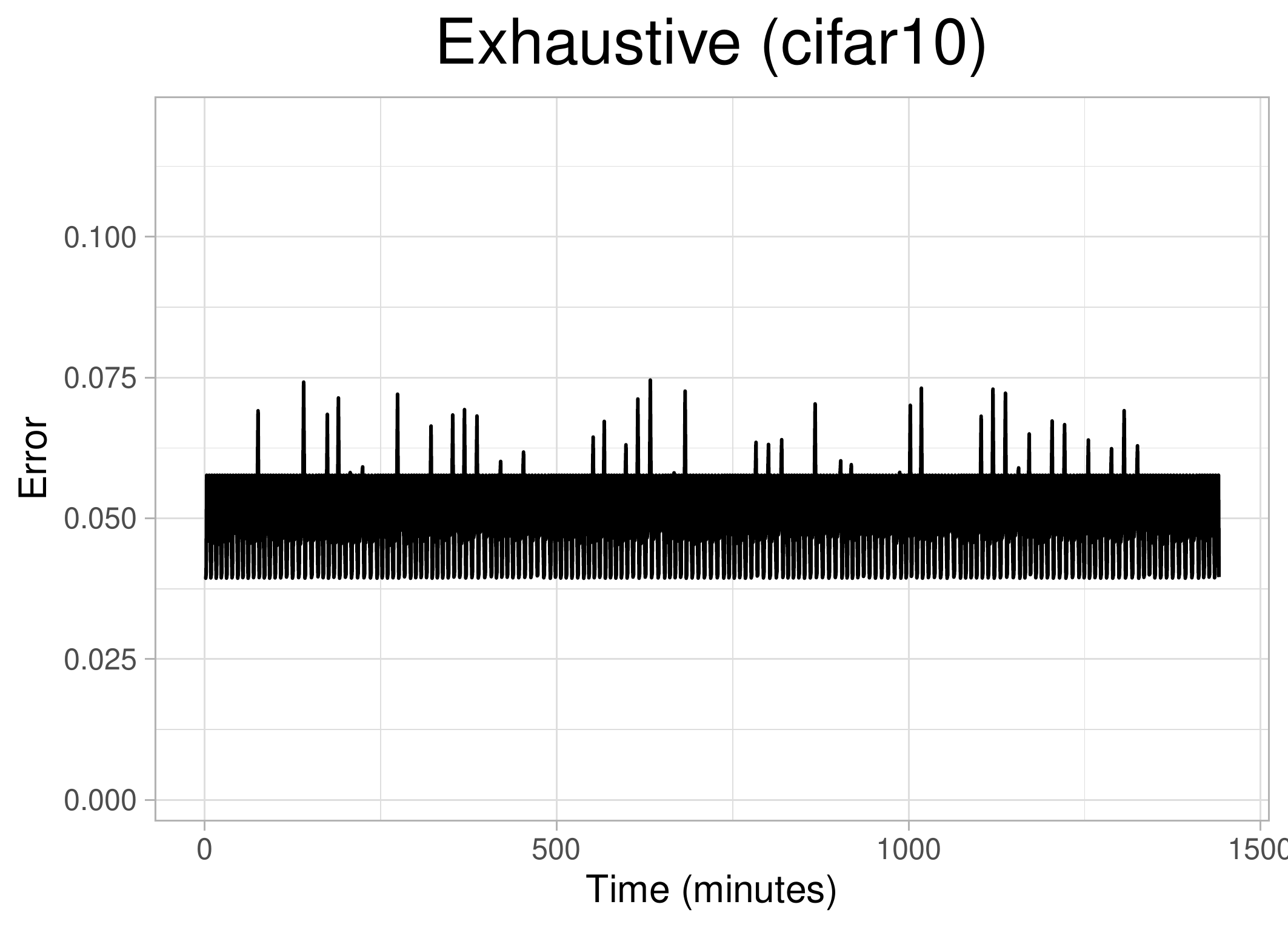} 
	\caption{\hl{Solutions explored through time}}
	\label{Fig.SolutionsTimeExhaustive1}
\end{figure}

\begin{figure}[h!]\centering
	\includegraphics[width=.5\linewidth]{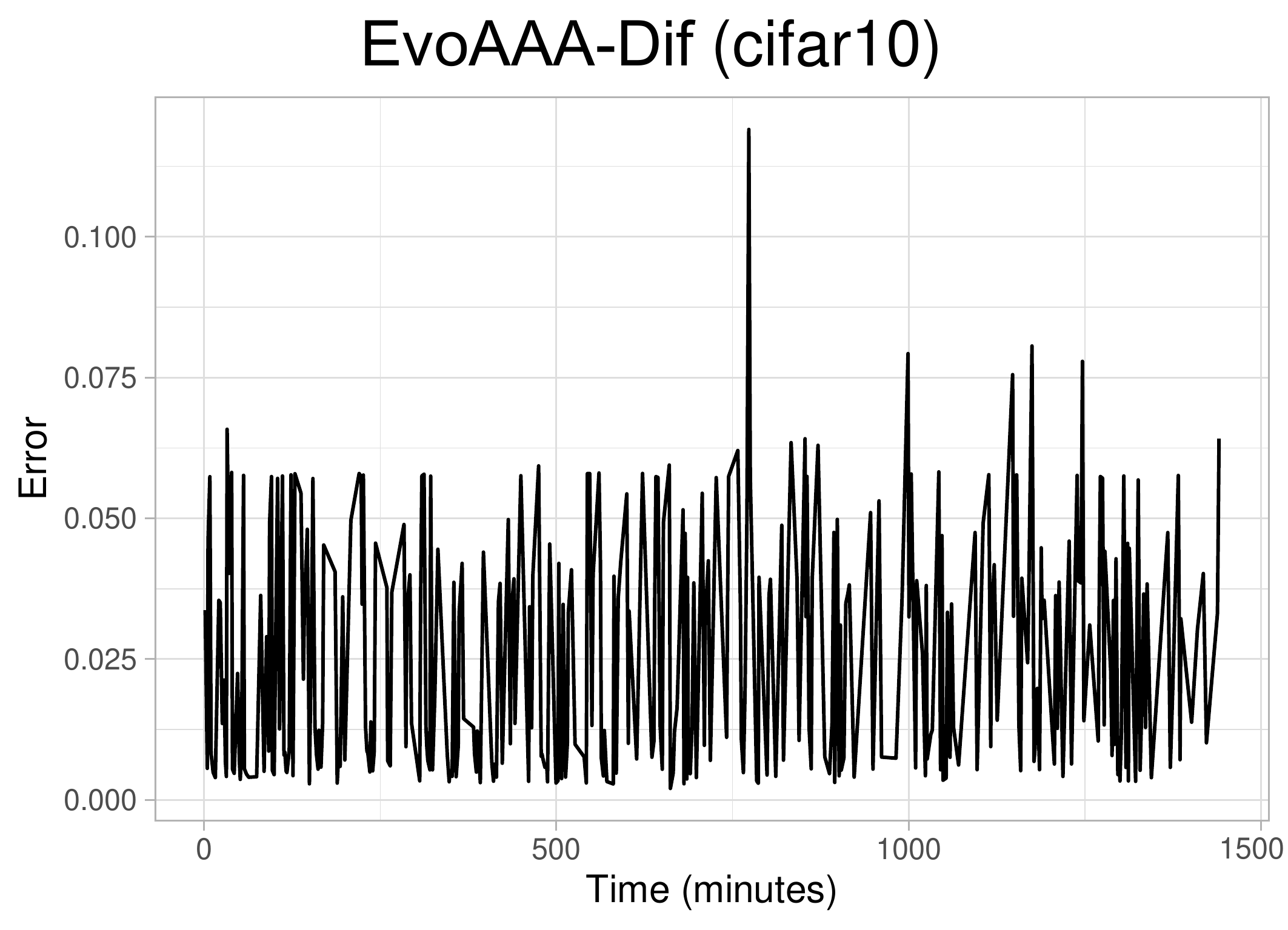} 
	\caption{\hl{Solutions explored through time}}
	\label{Fig.SolutionsTimeDE}
\end{figure}

The following aspect to be analyzed is how each approach explores the solution space. For doing so, the plots in Figure~\ref{Fig.SolutionsTimeExhaustive1} \hl{to Figure~\ref{Fig.SolutionsTimeDE}} show the MSE (Y axis) of the solutions examined through time (X axis)\footnote{The analysis described here has been made with the overall results, although only some illustrative cases are graphically shown. Plots corresponding to all combinations of dataset $\times$ search method are available in the repository.}. Since not all methods take the same running time, there are differences among the X axis for a given data set.

\begin{figure}[h!]\centering
	\includegraphics[width=.5\linewidth]{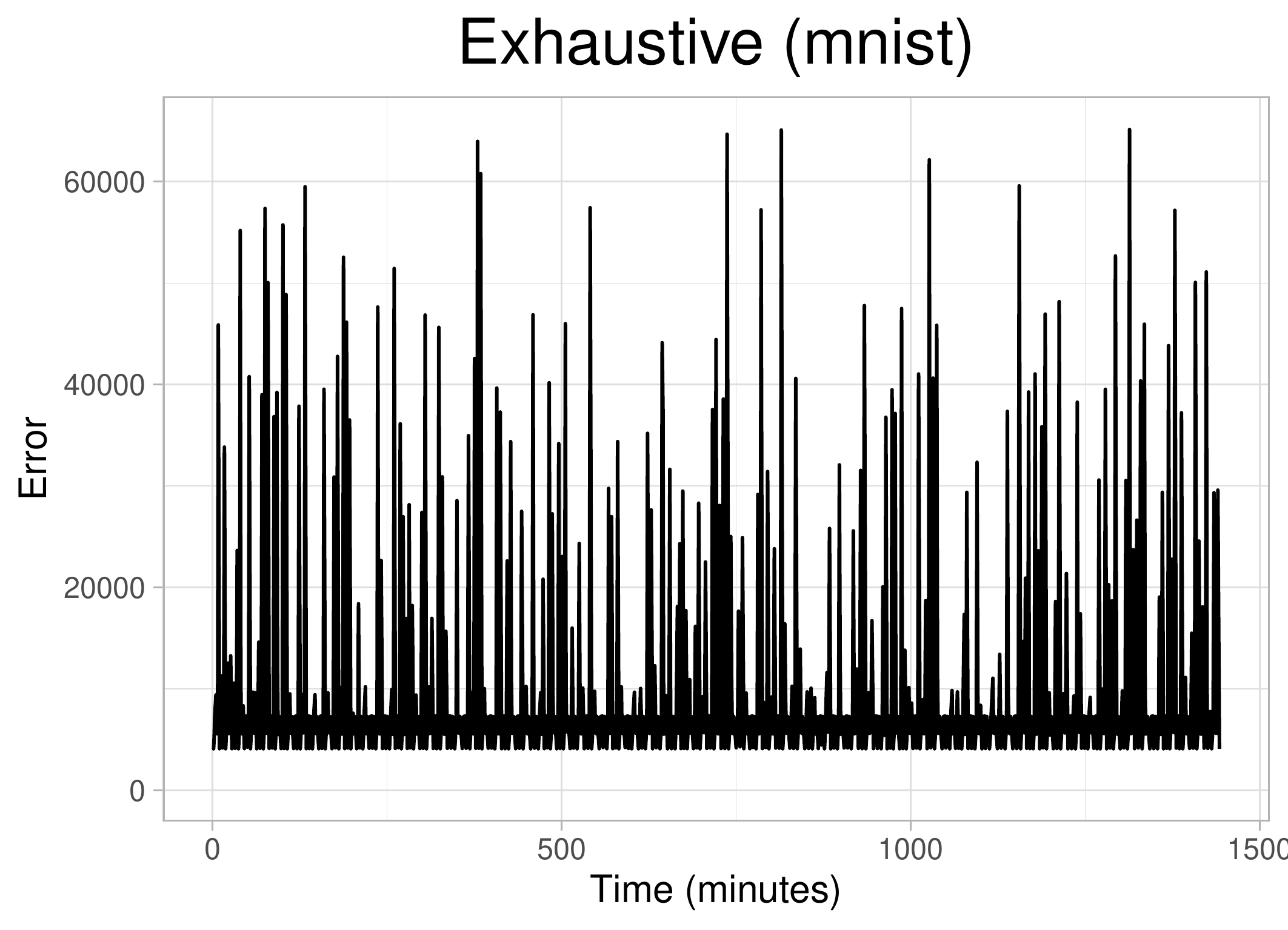} 
	\caption{\hl{Solutions explored through time}}
	\label{Fig.SolutionsTimeExhaustive2}
\end{figure}

As can be seen, the exhaustive search \hl{(Figure~\ref{Fig.SolutionsTimeExhaustive1})} keeps a high error rate from start to end. For fashion and mnist, whose attributes are quite similar, good and bad solutions are mixed over time \hl{(see Figure~\ref{Fig.SolutionsTimeExhaustive2})}. For the remainder data sets the search does not seem able to improve much as new solutions are evaluated.

\begin{figure}[h!]\centering
	\includegraphics[width=.5\linewidth]{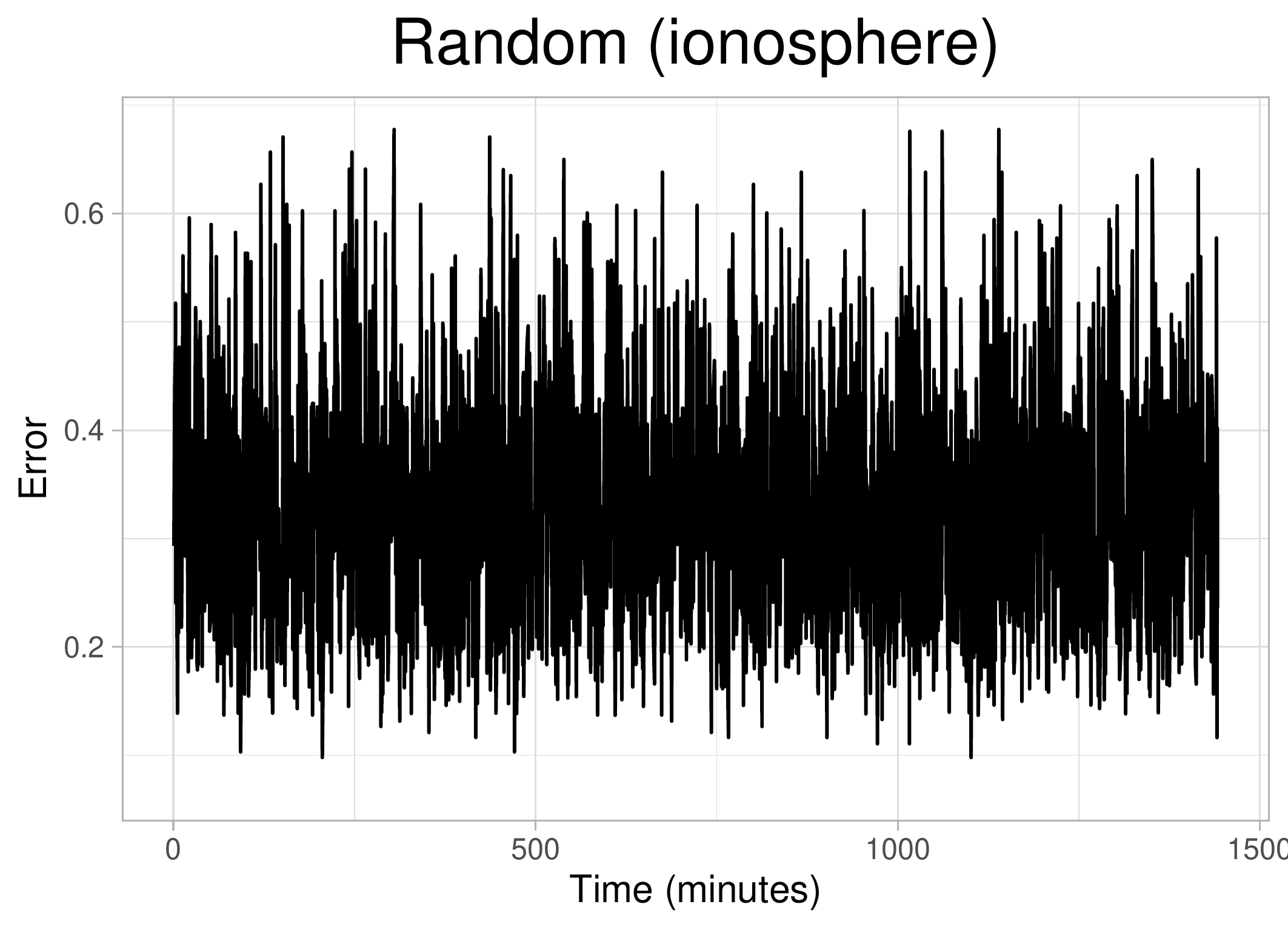} 
	\caption{\hl{Solutions explored through time}}
	\label{Fig.SolutionsTimeRandom}
\end{figure}

\hl{As can be observed in Figure~\ref{Fig.SolutionsTimeRandom}, the random search approach has a similar behavior to the exhaustive search, although it has larger fitness variability (the lines are less condensed) among the explored individuals. It is due to its ability to jump over all the solution space, instead of going through every possible combination of parameters. }

\begin{figure}[h!]\centering
	\includegraphics[width=.5\linewidth]{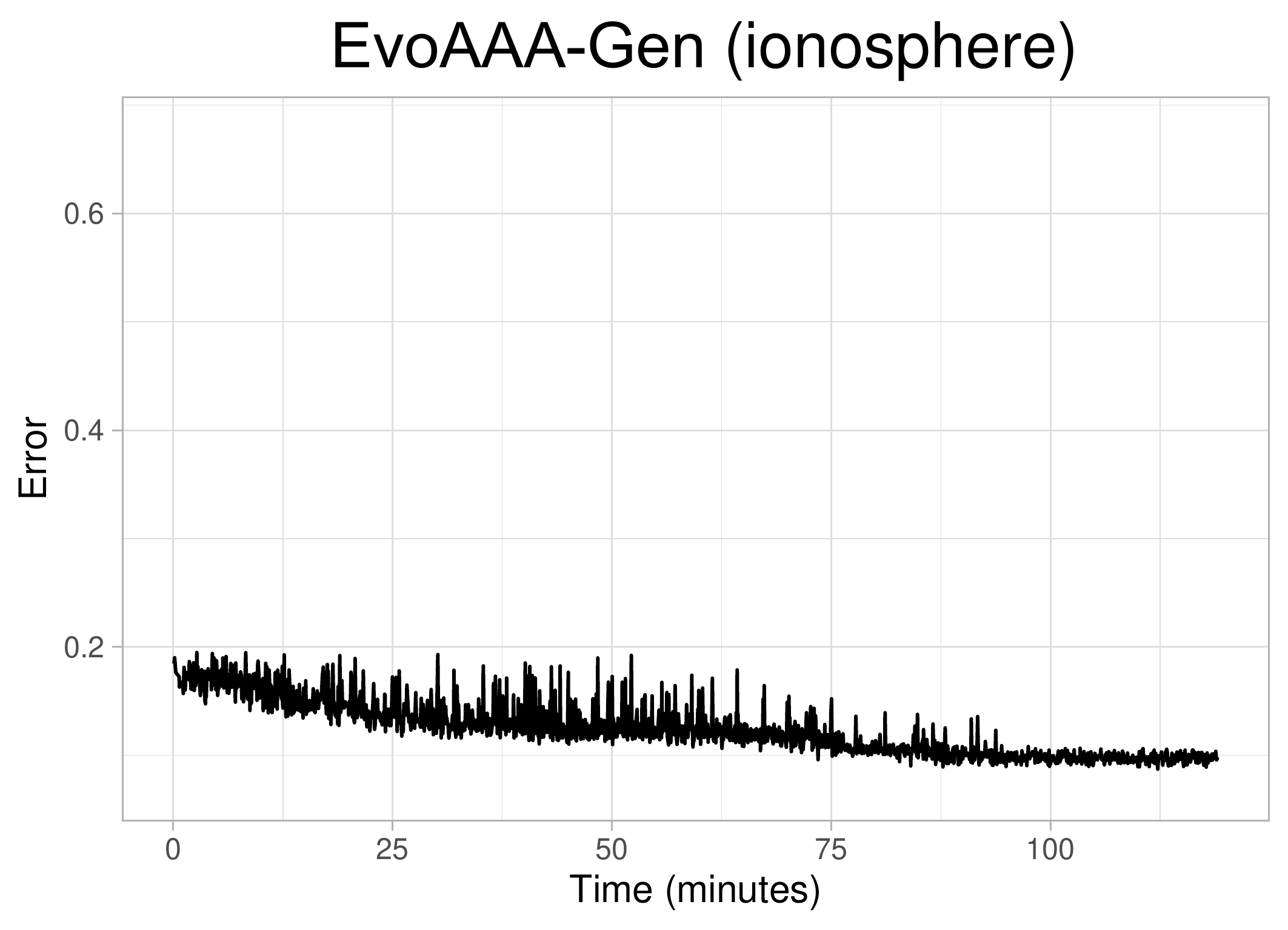} 
	\caption{\hl{Solutions explored through time}}
	\label{Fig.SolutionsTimeGA1}
\end{figure}

\begin{figure}[h!]\centering
	\includegraphics[width=.5\linewidth]{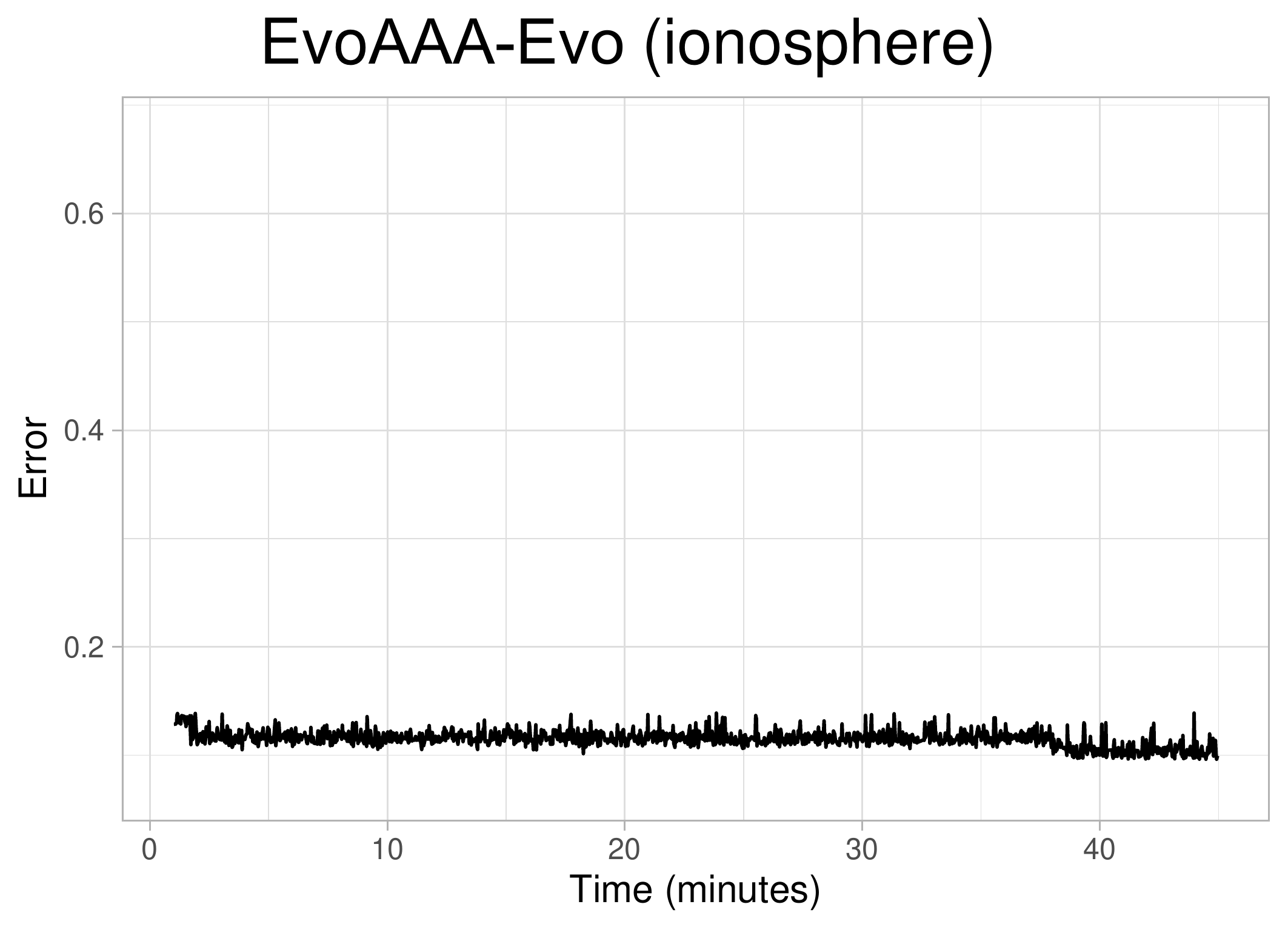} 
	\caption{\hl{Solutions explored through time}}
	\label{Fig.SolutionsTimeES1}
\end{figure}

The behavior of the \hl{{EvoAAA-Gen} and {EvoAAA-Evo}} candidates is similar\hl{, as shown in Figure~\ref{Fig.SolutionsTimeGA1} and Figure~\ref{Fig.SolutionsTimeES1}}. Both start with lower error rates than the exhaustive \hl{and random strategies}, and then improve as the running time goes by. This advance in the quality of solutions is not highlighted in the plots, as the Y axis is kept fixed to ease the comparison among the \hl{five} approaches. 

\begin{figure}[h!]\centering
	\includegraphics[width=.5\linewidth]{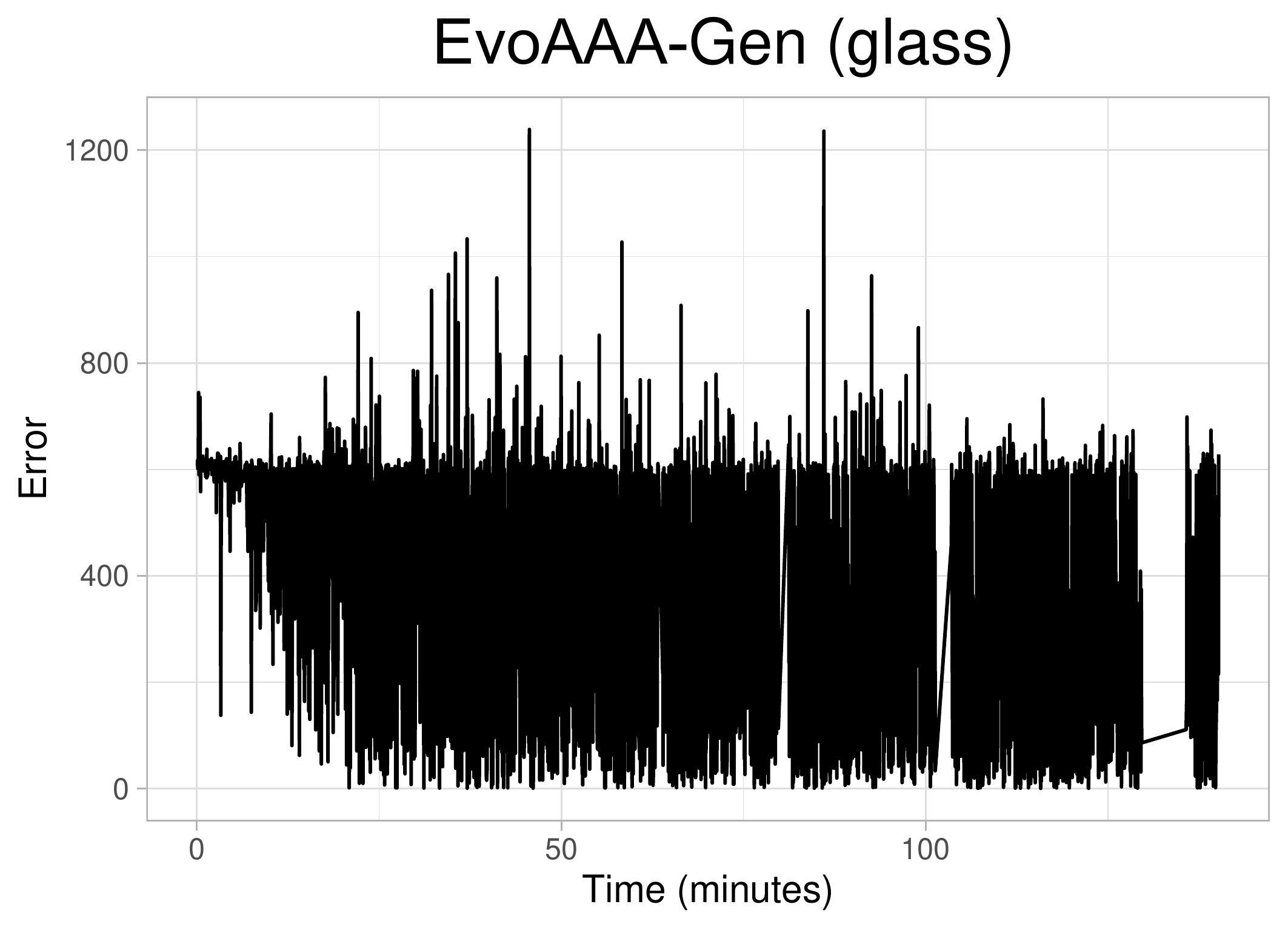} 
	\caption{\hl{Solutions explored through time}}
	\label{Fig.SolutionsTimeGA2}
\end{figure}

\begin{figure}[h!]\centering
	\includegraphics[width=.5\linewidth]{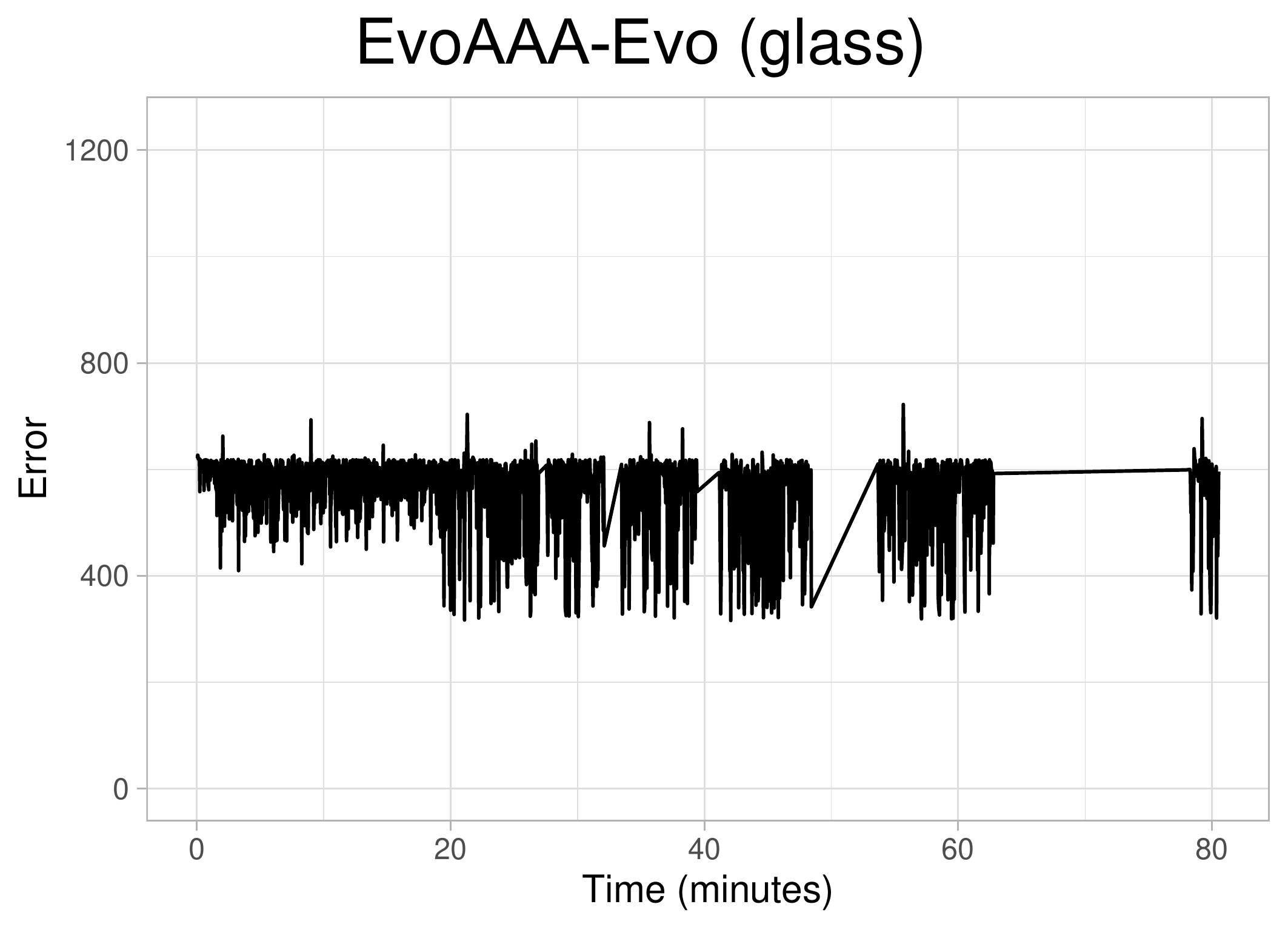} 
	\caption{\hl{Solutions explored through time}}
	\label{Fig.SolutionsTimeES2}
\end{figure}

The behavior shown by the {EvoAAA-Evo} and {EvoAAA-Gen} methods while working with the glass data set is anomalous as can be observed in Figure~\ref{Fig.SolutionsTimeGA2} and Figure~\ref{Fig.SolutionsTimeES2}. As stated in Table~\ref{Table.Datasets}, this is a data set with only 9 attributes and a handful of instances. It is probably the hardest case for an AE, since there is no much room to reduce the data representation nor enough patterns to learn it.

\hl{By contrast with {EvoAAA-Gen} and {EvoAAA-Evo}, the {EvoAAA-Dif} algorithm seems to have a wider exploration range in most cases, probably due to its larger population. Figure~\ref{Fig.SolutionsTimeDE} shows the quality of individuals evaluated through time by this search approach.}

\subsubsection{Achieved improvement vs explored solutions}

\begin{figure}[ht!]\centering
	\includegraphics[width=.5\linewidth]{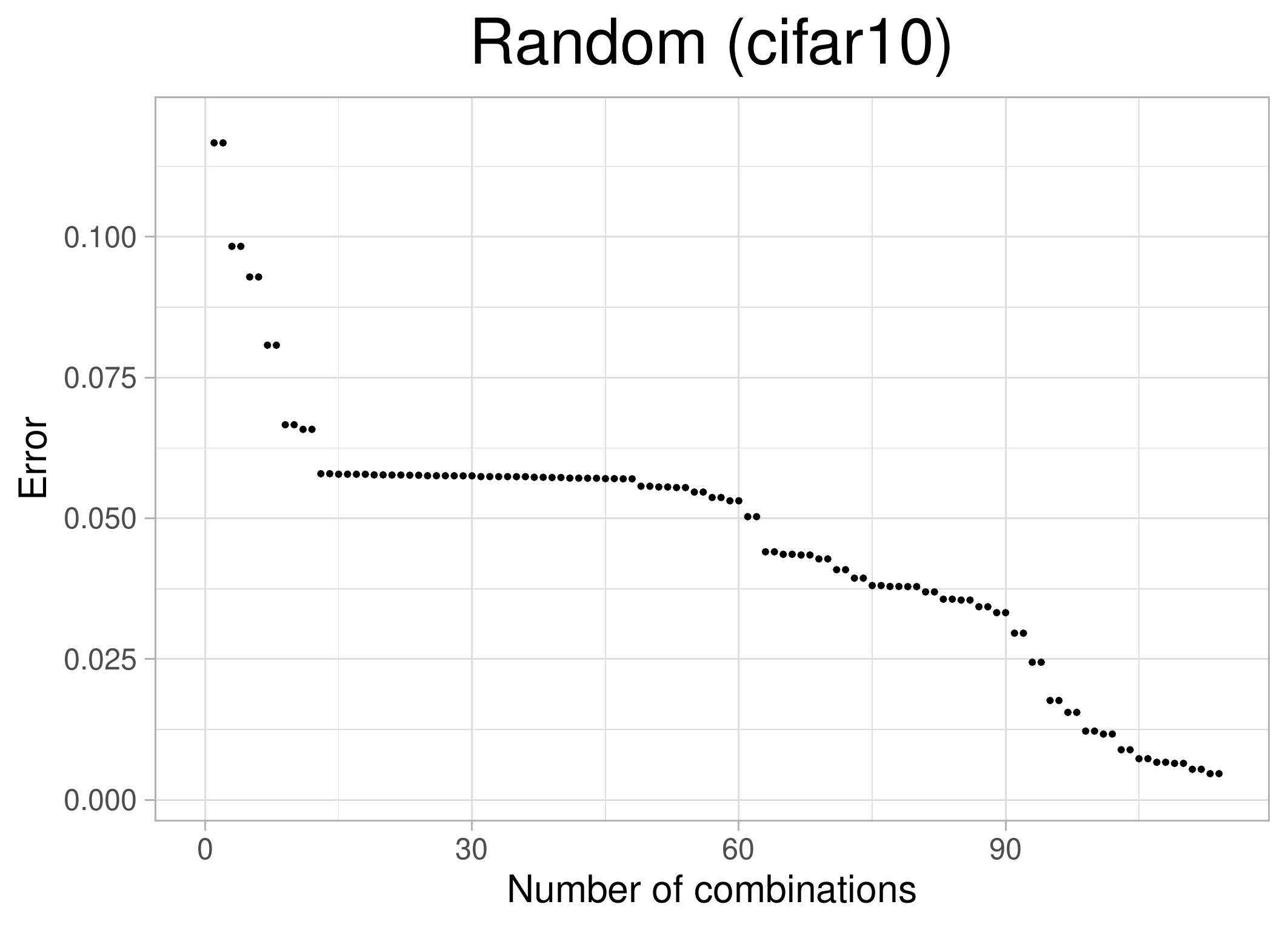} 
	\caption{\hl{Improvement achieved as solutions are explored}}
	\label{Fig.BestSolutions1}
\end{figure}

\begin{figure}[ht!]\centering
	\includegraphics[width=.5\linewidth]{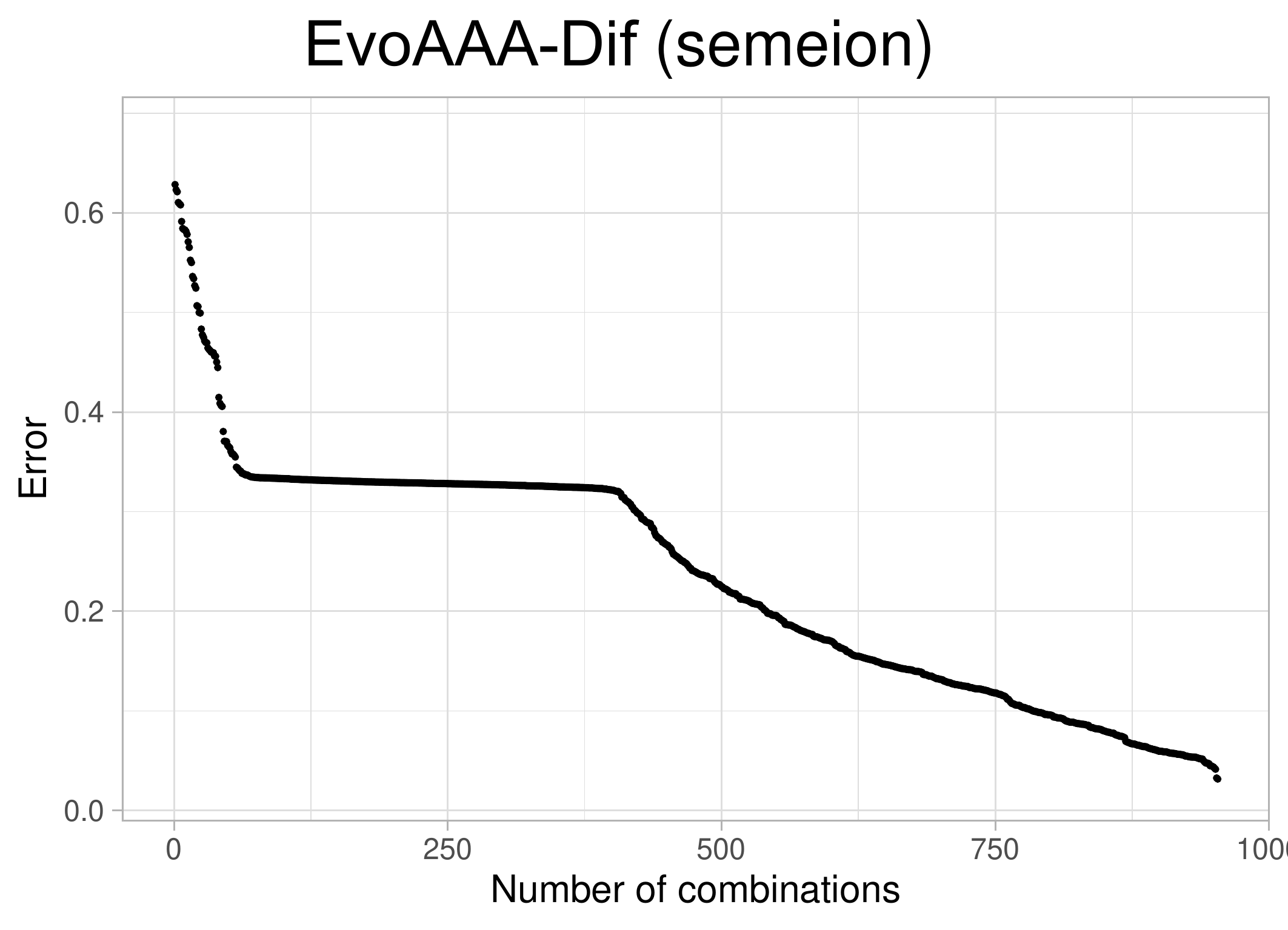} 
	\caption{\hl{Improvement achieved as solutions are explored}}
	\label{Fig.BestSolutions3}
\end{figure}

In Figures~\ref{Fig.BestSolutions1} \hl{to \ref{Fig.BestSolutions3}} the X axis has been changed from time to number of explored solutions, while the Y axis shows the error of the best solution found until now. The Y axis scale is kept fixed for each dataset, but the X axis scale changes since each approach examines a different amount of candidates. The goal is to analyze the improvement achieved by each optimization strategy as the they explore more solution space. \hl{All plots are available in the repository.}

As might be expected, all \hl{five} methods find better solutions as the number of possible architectures examined grows. However, \hl{both} the exhaustive method \hl{and random search} show some rungs as the search progresses, getting sometimes stuck for a long time in the same error level. \hl{An example of this behavior can be seen in Figure~\ref{Fig.BestSolutions1}}. 

\begin{figure}[ht!]\centering
	\includegraphics[width=.5\linewidth]{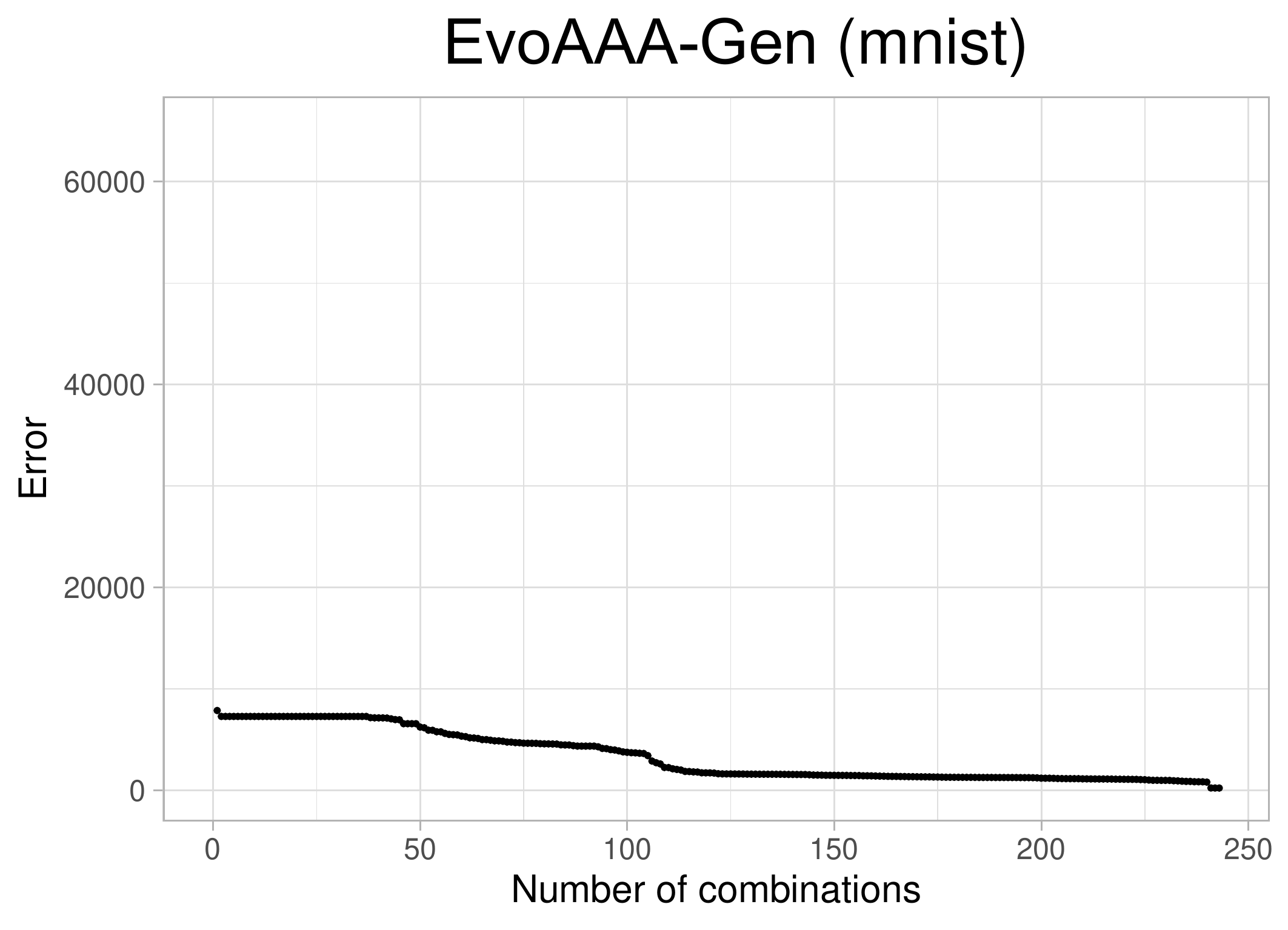} 
	\caption{\hl{Improvement achieved as solutions are explored}}
	\label{Fig.BestSolutions2}
\end{figure}

The improvements achieved by \hl{{EvoAAA-Gen} and {EvoAAA-Evo}} seem more progressive until they reach their minimum level \hl{(see Figure~\ref{Fig.BestSolutions2})}. In addition, this lower error rate is achieved after exploring a lower number of potential solutions. 

\hl{The behavior of the {EvoAAA-Dif} algorithm is somehow a mix of the previous ones, with a continuous improvement of results and starting with a lower error but having certain similarities with the random strategy.}

\subsubsection{Search convergence}

\hl{The following step is to analyze} the methods' speed of convergence. In this case there are nine plots \hl{available in the repository}, one per data set. \hl{One of them, shown in Figure~\ref{Fig.Convergence}, is taken as reference for the following analysis.} This way the same X and Y scales are shared by the \hl{five optimization strategies}. The X axis corresponds to running time. Only a portion of the time spent is represented, otherwise the lines that depict the EvoAAA instances would occupy only a small portion of the area, to the left. \hl{The lines that continue to the right mean that better values were found later, whereas those that do not reach the X limit indicate the best value achieved is in the plot (i.e. random search with the mnist dataset).} From these plots the following conclusions can be extracted:

\begin{figure}[ht!]\centering
	\includegraphics[width=.5\linewidth]{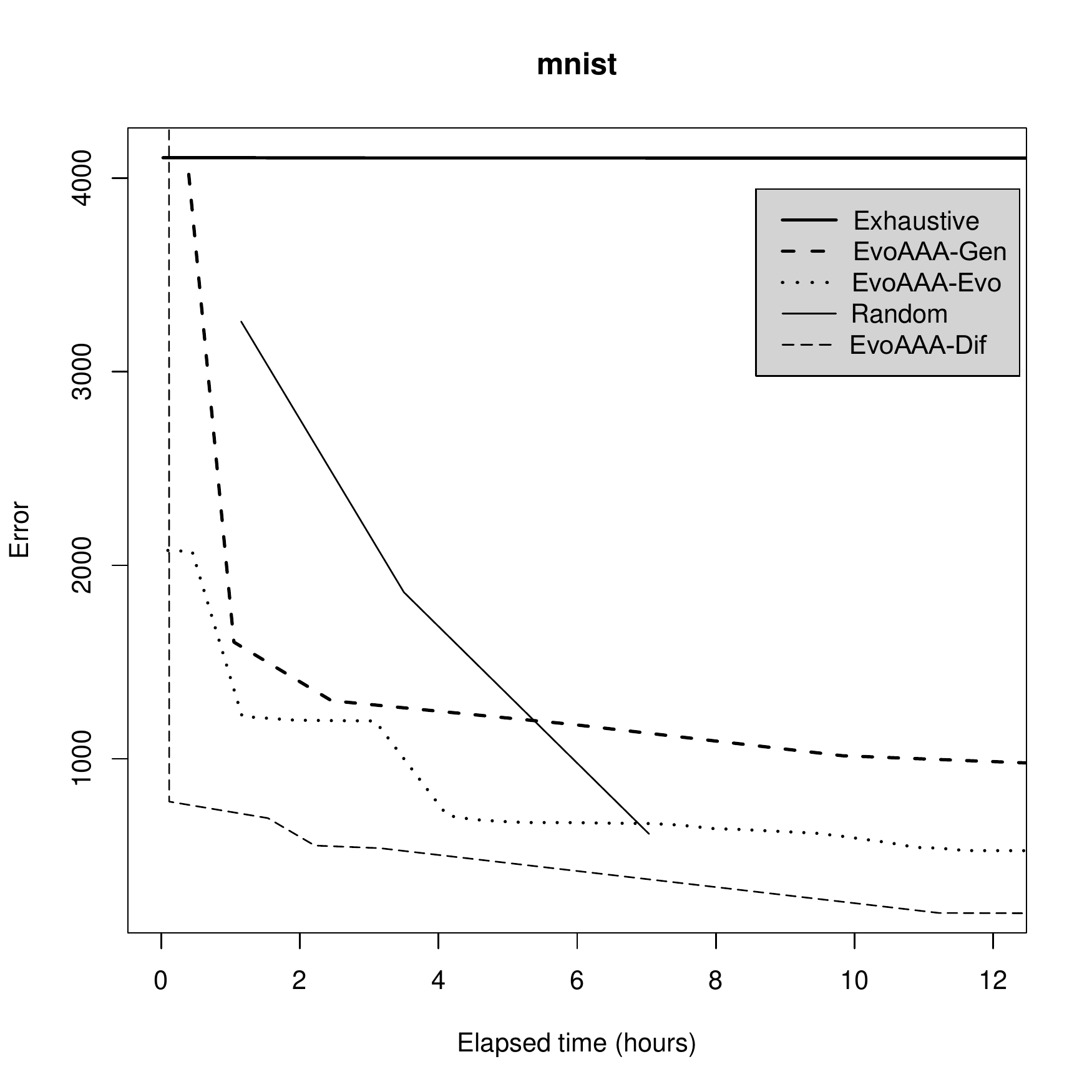} 
	\caption{\hl{Speed of convergence for each EvoAAA approach against exhaustive and random search}}
	\label{Fig.Convergence}
\end{figure}

\begin{itemize}
    \item In general, the exhaustive approach \hl{(thick solid line)} is stuck in a high error rate most of the time. The only exception is the glass data set, with this search method almost chasing the \hl{three EvoAAA} strategies.
    
    \hl{\item The random search (thin solid line) behaves aimlessly as it would be expected. Sometimes it shows a slow progress over time similar to that of the exhaustive strategy (e.g. cifar10 and delicious) while other times it finds a good solution very quickly and then no better solutions are found (e.g. fashion and mnist). It is a strategy that could provide a good result in short time but without guarantees.}
    
    \item \hl{In general, }the \hl{three} EvoAAA instances converge quickly to a lower error value \hl{than the two baseline strategies}, then stabilize and keep improving at a slower rate.
    
    \item In most occasions the ES approach (dotted line) reaches its optimum (lowest error value) before the GA (thick dashed line) does, then finishes the execution. On the contrary, the GA method keeps improving for longer. Due to this behavior, it is able to beat ES in many cases.
    
    \hl{\item The DE strategy (thin dashed line) usually starts with worse solutions than GA and ES, but in most cases it converges faster, particularly with the most complex datasets such as cifar10, fashion and mnist. It is able to get the best result in a fraction of the time with respect to the other alternatives in some cases.}
\end{itemize}

On the basis of this analysis, it would be possible to adjust the parameters of the \hl{DE and }ES algorithms, increasing the population or \hl{their} number of iterations, in order to run longer and keep improving as the GA does. \hl{They} would presumably achieve results comparable to those of the GA method \hl{for some small datasets (e.g. sonar, glass and spect), or even better with DE}.

\hl{
\subsubsection{Influence of the penalization factor}\label{Sec.PenalizationAnalysis}
To finish this analysis, how the penalization factor linked to the AEs complexity influence the obtained results is scrutinized. For doing so, the DE strategy has been used over the sonar dataset with $\alpha$ varying from 1 to 0 following a logarithmic scale.

The goal of the $\alpha$ penalization factor is to prefer simpler AE architectures for similar reconstruction performances. Intuitively, lower $\alpha$ values would produce AEs with a higher reconstruction power but also with more layers and units, and the opposite for higher $\alpha$ values (i.e. simpler architectures having lower reconstruction accuracy).
}

\begin{figure}[h!]\centering
  \includegraphics[width=.5\linewidth]{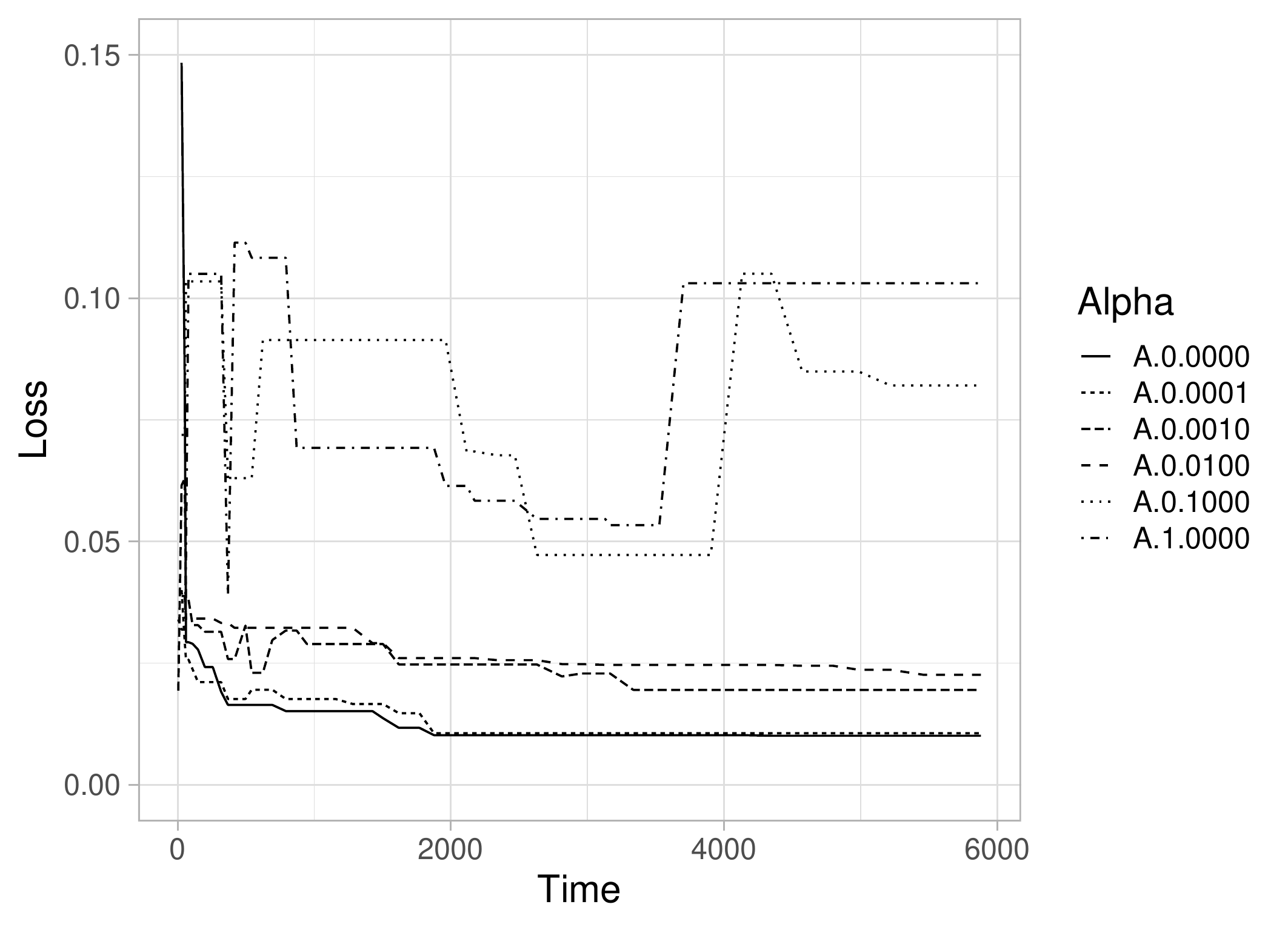}
  \caption{\hl{Loss change through time for different $\alpha$ values  using EvoAAA-Dif with the sonar dataset}}
  \label{Fig.AlphaLoss}
\end{figure}

\hl{To start with this analysis, Figure~\ref{Fig.AlphaLoss} shows how the MSE changes through time, as the architectures are explored by the DE algorithm, for the considered $\alpha$ values. As can be observed, the two highest penalization values severely affect the search procedure as the abrupt loss changes reflect. As the DE method tries to improve the performance testing more complex DEs the penalization grows. These rungs are reduced as the $\alpha$ value lowers, until there is no impact with $\alpha = 0$. So, the first outcome would be that it is preferable to have small penalization factors, specifically values that are a fraction of the MSE.}

\begin{figure}[h!]
	\centering
  \includegraphics[width=.5\linewidth]{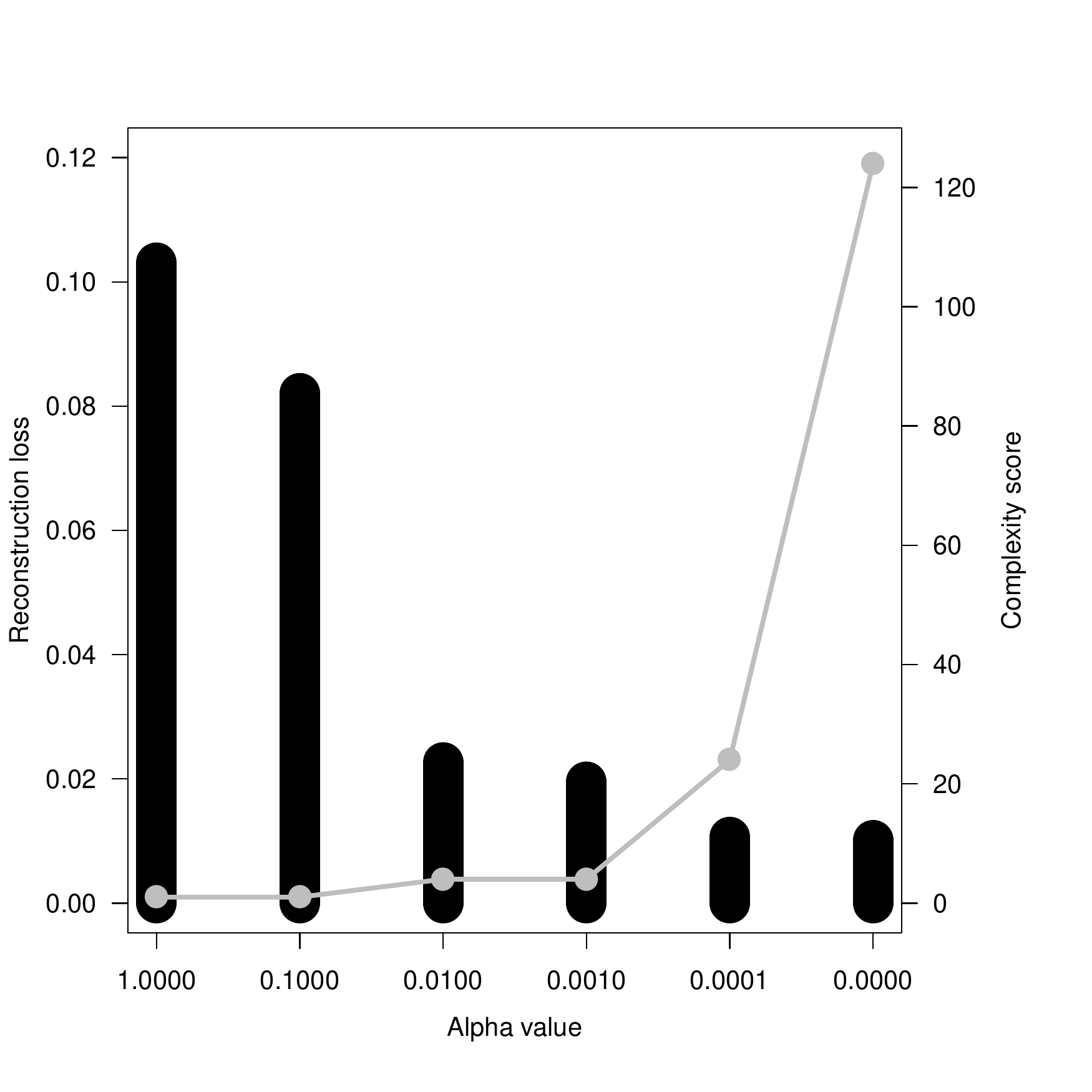}
  \caption{\hl{Loss vs AE complexity for different $\alpha$ values using EvoAAA-Dif with the sonar dataset}}
  \label{Fig.LossComplexity}
\end{figure}

\hl{In order to better appreciate the extent to which performance degrades as $\alpha$ increases, the final results obtained with DE from the sonar dataset for each penalization have been represented in Figure~\ref{Fig.LossComplexity}. Black bars are linked to the left Y axis scale and denote MSE (higher bars are worse), while the line indicates complexity level (right Y axis). Observe that for $\alpha = 0$ and $\alpha = 0.0001$ there is almost no change in performance, but the difference in complexity is remarkable. As penalization factor increases, to the left, the complexity scores reduces but the reconstruction error does the opposite.}

\begin{table}[h!]
\centering
\renewcommand{\arraystretch}{0.9}
\hl{
\caption{Loss and AE configuration obtained with each $\alpha$ value  using EvoAAA-Dif with the sonar dataset}
\label{Table.Alphas}
\begin{tabular}{ccc}
  \toprule
  \textbf{$\alpha$} & \textbf{Loss} & \textbf{Layers and units} \\ 
  \midrule
  0.0000 & 0.00993972 & 57, 47, 36, 22, 36, 47, 57 \\
  0.0001 & 0.01062195 & 37, 32, 8, 32, 37 \\
  0.0010 & 0.01951711 & 23, 2, 23 \\
  0.0100 & 0.02262429 & 38, 3, 38 \\
  0.1000 & 0.08206710 & 1 \\
  1.0000 & 0.10308630 & 1 \\
  \bottomrule 
\end{tabular}
}
\end{table}

\hl{The exact MSE values and configuration for each considered $\alpha$ are summarized in Table~\ref{Table.Alphas}. As indicated above, the performance difference between $\alpha = 0$ and $\alpha = 0.0001$ is almost negligible, but the former uses 7 layers instead of 5 for the latter, and the encoding length is 22 units against only 8. Clearly, in this specific case a small penalization factor returns better AE architectures than higher ones or having no penalization at all.

As a result of this analysis, $\alpha$ was set to 0.0001 for the experiments conducted in this study, as was specified at the beginning of Section~\ref{Sec.Experiments}.
}

\section{Concluding remarks}\label{Sec.Conclusions}

AEs are a useful tool to \hl{face} representation learning. \hl{However, finding the } AE architecture \hl{that fits better} every case is a difficult task. Internal cross-validation is an usual approach for tuning hyperparameters. However, the solution space is huge \hl{if we want to also adjust the AE architecture}. Therefore, more powerful \hl{methods} to face this problem would be needed.

In this paper we have proposed EvoAAA, an evolutionary based approach to find the best AE architecture for each data set. First, a way of encoding the AE architecture within a chromosome has been proposed. It is broad enough to consider most AE variations, including different amounts of layers and units, individual activation functions per layer and several loss functions. Then, \hl{three} search methods have been planned, one based on \hl{differential evolution, another one founded on }a genetic algorithm and the other on an evolutionary strategy. Lastly, a thorough experimentation and analysis have been conducted\hl{, comparing the results of the three EvoAAA strategies against two different baselines, exhaustive and random search}.

Overall, it has been demonstrated that the proposed methodology is able to find a good AE structure for each data set. It may not be the optimal one, \hl{as more advanced search algorithms and optimization strategies could improve these results, but it is better than the ones randomly chosen or found through an exhaustive look up if results in a reasonable time are needed}. The conducted experiments demonstrate that EvoAAA is a competitive procedure to accomplish the job.

\begin{quotation}
	This work was partially supported by the project TIN2015-68854-R (FEDER Founds) of the
	Spanish Ministry of Economy and Competitiveness. 
\end{quotation}



\nocite{*} 

\begin{thebibliography}{90}
	\ifx \bisbn   \undefined \def \bisbn  #1{ISBN #1}\fi
	\ifx \binits  \undefined \def \binits#1{#1} \fi
	\ifx \bauthor  \undefined \def \bauthor#1{#1} \fi
	\ifx \bjtitle  \undefined \def \bjtitle#1{\textit{#1}}\fi
	\ifx \batitle  \undefined \def \batitle#1{#1} \fi
	\ifx \bctitle  \undefined \def \bctitle#1{#1} \fi
	\ifx \bvolume  \undefined \def \bvolume#1{\textbf{#1}}\fi
	\ifx \byear  \undefined \def \byear#1{#1} \fi
	\ifx \bissue  \undefined \def \bissue#1{#1} \fi
	\ifx \bfpage  \undefined \def \bfpage#1{#1} \fi
	\ifx \blpage  \undefined \def \blpage #1{#1} \fi
	\ifx \burl  \undefined \def \burl#1{#1} \fi
	\ifx \doiurl  \undefined \def \doiurl#1{#1} \fi
	\ifx \betal  \undefined \def \betal{et al.} \fi
	\ifx \binstitute  \undefined \def \binstitute#1{#1} \fi
	\ifx \beditor  \undefined \def \beditor#1{#1} \fi
	\ifx \bpublisher  \undefined \def \bpublisher#1{#1} \fi
	\ifx \bbtitle  \undefined \def \bbtitle#1{\textit{#1}} \fi
	\ifx \bedition  \undefined \def \bedition#1{#1} \fi
	\ifx \bseriesno  \undefined \def \bseriesno#1{#1} \fi
	\ifx \blocation  \undefined \def \blocation#1{#1} \fi
	\ifx \bsertitle  \undefined \def \bsertitle#1{#1} \fi
	\ifx \bsnm \undefined \def \bsnm#1{#1} \fi
	\ifx \bsuffix \undefined \def \bsuffix#1{#1} \fi
	\ifx \bparticle \undefined \def \bparticle#1{#1} \fi
	\ifx \barticle \undefined \def \barticle#1{#1} \fi
	\ifx \botherref \undefined \def \botherref #1{#1} \fi
	\ifx \url \undefined \def \url#1{#1} \fi
	\ifx \bchapter \undefined \def \bchapter#1{#1} \fi
	\ifx \bbook \undefined \def \bbook#1{#1} \fi
	\ifx \bcomment \undefined \def \bcomment#1{#1} \fi
	\ifx \oauthor \undefined \def \oauthor#1{#1} \fi
	\ifx \citeauthoryear \undefined \def \citeauthoryear#1{#1} \fi
	\ifx \texttildelow  \undefined \def \texttildelow{\symbol{126}} \fi
	\def \endbibitem {}
	\ifx \bconflocation  \undefined \def \bconflocation#1{#1} \fi
	
	\bibitem{bishop2006pattern}
	\begin{bbook}
		\bauthor{\binits{C.M.}~\bsnm{Bishop}},
		\bbtitle{Pattern recognition and machine learning},
		\bpublisher{Springer},
		\byear{2006}.
	\end{bbook}
	\endbibitem
	
	\bibitem{guzella2009review}
	\begin{barticle}
		\bauthor{\binits{T.S.}~\bsnm{Guzella}} and
		\bauthor{\binits{W.M.}~\bsnm{Caminhas}},
		\batitle{A review of machine learning approaches to spam filtering},
		\bjtitle{Expert Systems with Applications}
		\bvolume{36}(\bissue{7})
		(\byear{2009}),
		\bfpage{10206}--\blpage{10222}.
	\end{barticle}
	\endbibitem
	
	\bibitem{bhattacharyya2011data}
	\begin{barticle}
		\bauthor{\binits{S.}~\bsnm{Bhattacharyya}},
		\bauthor{\binits{S.}~\bsnm{Jha}},
		\bauthor{\binits{K.}~\bsnm{Tharakunnel}} and
		\bauthor{\binits{J.C.}~\bsnm{Westland}},
		\batitle{Data mining for credit card fraud: A comparative study},
		\bjtitle{Decision Support Systems}
		\bvolume{50}(\bissue{3})
		(\byear{2011}),
		\bfpage{602}--\blpage{613}.
	\end{barticle}
	\endbibitem
	
	\bibitem{schafer1999recommender}
	\begin{bchapter}
		\bauthor{\binits{J.B.}~\bsnm{Schafer}},
		\bauthor{\binits{J.}~\bsnm{Konstan}} and
		\bauthor{\binits{J.}~\bsnm{Riedl}},
		\bctitle{Recommender systems in e-commerce},
		in: \bbtitle{Proceedings of the 1st ACM conference on Electronic commerce},
		\binstitute{ACM},
		\byear{1999},
		pp.~\bfpage{158}--\blpage{166}.
	\end{bchapter}
	\endbibitem
	
	\bibitem{Domingos2012AFU}
	\begin{barticle}
		\bauthor{\binits{P.}~\bsnm{Domingos}},
		\batitle{A few useful things to know about machine learning},
		\bjtitle{Communications of the ACM}
		\bvolume{55}(\bissue{10})
		(\byear{2012}),
		\bfpage{78}--\blpage{87}.
	\end{barticle}
	\endbibitem
	
	\bibitem{DataPreprocessing}
	\begin{bbook}
		\bauthor{\binits{S.}~\bsnm{Garc{\'\i}a}},
		\bauthor{\binits{J.}~\bsnm{Luengo}} and
		\bauthor{\binits{F.}~\bsnm{Herrera}},
		\bbtitle{Data preprocessing in data mining},
		\bpublisher{Springer},
		\byear{2015},
		pp.~\bfpage{163}--\blpage{194},
		\bcomment{Chapter~7}.
		ISBN \bisbn{978-3-319-10247-4}.
	\end{bbook}
	\endbibitem
	
	\bibitem{CorrelationFS}
	\begin{botherref}
		\oauthor{\binits{M.A.}~\bsnm{Hall}},
		Correlation-based feature selection for machine learning,
		PhD thesis,
		University of Waikato Hamilton,
		1999.
	\end{botherref}
	\endbibitem
	
	\bibitem{MutualInformationDS}
	\begin{barticle}
		\bauthor{\binits{H.}~\bsnm{Peng}},
		\bauthor{\binits{F.}~\bsnm{Long}} and
		\bauthor{\binits{C.H.Q.}~\bsnm{Ding}},
		\batitle{Feature selection based on mutual information criteria of
			max-dependency, max-relevance, and min-redundancy},
		\bjtitle{IEEE Transactions on Pattern Analysis and Machine Intelligence}
		\bvolume{27}
		(\byear{2005}),
		\bfpage{1226}--\blpage{1238}.
	\end{barticle}
	\endbibitem
	
	\bibitem{FeatureExtractionIntro}
	\begin{bchapter}
		\bauthor{\binits{I.}~\bsnm{Guyon}} and
		\bauthor{\binits{A.}~\bsnm{Elisseeff}},
		\bbtitle{An Introduction to Feature Extraction},
		in: \bbtitle{Feature Extraction: Foundations and Applications},
		\beditor{\binits{I.}~\bsnm{Guyon}},
		\beditor{\binits{M.}~\bsnm{Nikravesh}},
		\beditor{\binits{S.}~\bsnm{Gunn}} and
		\beditor{\binits{L.A.}~\bsnm{Zadeh}}, eds,
		\bpublisher{Springer Berlin Heidelberg},
		\blocation{Berlin, Heidelberg},
		\byear{2006},
		pp.~\bfpage{1}--\blpage{25}.
	\end{bchapter}
	\endbibitem
	
	\bibitem{rumelhart1988learning}
	\begin{barticle}
		\bauthor{\binits{D.E.}~\bsnm{Rumelhart}},
		\bauthor{\binits{G.E.}~\bsnm{Hinton}},
		\bauthor{\binits{R.J.}~\bsnm{Williams}} \betal,
		\batitle{Learning representations by back-propagating errors},
		\bjtitle{Cognitive modeling}
		\bvolume{5}(\bissue{3})
		(\byear{1988}),
		\bfpage{1}.
	\end{barticle}
	\endbibitem
	
	\bibitem{bengio_representation_2013}
	\begin{barticle}
		\bauthor{\binits{Y.}~\bsnm{Bengio}},
		\bauthor{\binits{A.}~\bsnm{Courville}} and
		\bauthor{\binits{P.}~\bsnm{Vincent}},
		\batitle{Representation learning: A review and new perspectives},
		\bjtitle{{IEEE} transactions on pattern analysis and machine intelligence}
		\bvolume{35}(\bissue{8})
		(\byear{2013}),
		\bfpage{1798}--\blpage{1828}.
	\end{barticle}
	\endbibitem
	
	\bibitem{lecun2015deep}
	\begin{barticle}
		\bauthor{\binits{Y.}~\bsnm{LeCun}},
		\bauthor{\binits{Y.}~\bsnm{Bengio}} and
		\bauthor{\binits{G.}~\bsnm{Hinton}},
		\batitle{Deep learning},
		\bjtitle{Nature}
		\bvolume{521}(\bissue{7553})
		(\byear{2015}),
		\bfpage{436}.
	\end{barticle}
	\endbibitem
	
	\bibitem{goodfellow2016deep}
	\begin{bbook}
		\bauthor{\binits{I.}~\bsnm{Goodfellow}},
		\bauthor{\binits{Y.}~\bsnm{Bengio}} and
		\bauthor{\binits{A.}~\bsnm{Courville}},
		\bbtitle{Deep learning},
		\bpublisher{MIT press},
		\byear{2016}.
	\end{bbook}
	\endbibitem
	
	\bibitem{hinton2006reducing}
	\begin{barticle}
		\bauthor{\binits{G.E.}~\bsnm{Hinton}} and
		\bauthor{\binits{R.R.}~\bsnm{Salakhutdinov}},
		\batitle{Reducing the dimensionality of data with neural networks},
		\bjtitle{Science}
		\bvolume{313}(\bissue{5786})
		(\byear{2006}),
		\bfpage{504}--\blpage{507}.
	\end{barticle}
	\endbibitem
	
	\bibitem{vincent2008extracting}
	\begin{bchapter}
		\bauthor{\binits{P.}~\bsnm{Vincent}},
		\bauthor{\binits{H.}~\bsnm{Larochelle}},
		\bauthor{\binits{Y.}~\bsnm{Bengio}} and
		\bauthor{\binits{P.-A.}~\bsnm{Manzagol}},
		\bctitle{Extracting and composing robust features with denoising autoencoders},
		in: \bbtitle{Proceedings of the 25th international conference on Machine
			learning},
		\binstitute{ACM},
		\byear{2008},
		pp.~\bfpage{1096}--\blpage{1103}.
	\end{bchapter}
	\endbibitem
	
	\bibitem{vincent2010stacked}
	\begin{barticle}
		\bauthor{\binits{P.}~\bsnm{Vincent}},
		\bauthor{\binits{H.}~\bsnm{Larochelle}},
		\bauthor{\binits{I.}~\bsnm{Lajoie}},
		\bauthor{\binits{Y.}~\bsnm{Bengio}} and
		\bauthor{\binits{P.-A.}~\bsnm{Manzagol}},
		\batitle{Stacked denoising autoencoders: Learning useful representations in a
			deep network with a local denoising criterion},
		\bjtitle{Journal of machine learning research}
		\bvolume{11}(\bissue{Dec})
		(\byear{2010}),
		\bfpage{3371}--\blpage{3408}.
	\end{barticle}
	\endbibitem
	
	\bibitem{Charte:ReviewAEs}
	\begin{barticle}
		\bauthor{\binits{D.}~\bsnm{Charte}},
		\bauthor{\binits{F.}~\bsnm{Charte}},
		\bauthor{\binits{S.}~\bsnm{Garc{\'\i}a}},
		\bauthor{\binits{M.J.}~\bsnm{del Jesus}} and
		\bauthor{\binits{F.}~\bsnm{Herrera}},
		\batitle{A practical tutorial on autoencoders for nonlinear feature fusion:
			Taxonomy, models, software and guidelines},
		\bjtitle{Information Fusion}
		\bvolume{44}
		(\byear{2018}),
		\bfpage{78}--\blpage{96}.
	\end{barticle}
	\endbibitem
	
	\bibitem{Charte:ShowcaseAEs}
	\begin{bchapter}
		\bauthor{\binits{D.}~\bsnm{Charte}},
		\bauthor{\binits{F.}~\bsnm{Charte}},
		\bauthor{\binits{M.J.}~\bsnm{del Jesus}} and
		\bauthor{\binits{F.}~\bsnm{Herrera}},
		\bctitle{A Showcase of the Use of Autoencoders in Feature Learning
			Applications},
		in: \bbtitle{From Bioinspired Systems and Biomedical Applications to Machine
			Learning},
		\beditor{\binits{J.M.}~\bsnm{Ferr{\'a}ndez~Vicente}},
		\beditor{\binits{J.R.}~\bsnm{{\'A}lvarez-S{\'a}nchez}},
		\beditor{\binits{F.}~\bsnm{de~la Paz~L{\'o}pez}},
		\beditor{\binits{J.}~\bsnm{Toledo~Moreo}} and
		\beditor{\binits{H.}~\bsnm{Adeli}}, eds,
		\bpublisher{Springer International Publishing},
		\byear{2019},
		pp.~\bfpage{412}--\blpage{421}.
		ISBN \bisbn{978-3-030-19651-6}.
	\end{bchapter}
	\endbibitem
	
	\bibitem{garey2002computers}
	\begin{bbook}
		\bauthor{\binits{M.R.}~\bsnm{Garey}} and
		\bauthor{\binits{D.S.}~\bsnm{Johnson}},
		\bbtitle{Computers and intractability},
		Vol.~\bseriesno{29},
		\bpublisher{wh freeman New York},
		\byear{2002}.
	\end{bbook}
	\endbibitem
	
	\bibitem{back1993overview}
	\begin{barticle}
		\bauthor{\binits{T.}~\bsnm{B{\"a}ck}} and
		\bauthor{\binits{H.-P.}~\bsnm{Schwefel}},
		\batitle{An overview of evolutionary algorithms for parameter optimization},
		\bjtitle{Evolutionary computation}
		\bvolume{1}(\bissue{1})
		(\byear{1993}),
		\bfpage{1}--\blpage{23}.
	\end{barticle}
	\endbibitem
	
	\bibitem{pulgar2020choosing}
	\begin{barticle}
		\bauthor{\binits{F.J.}~\bsnm{Pulgar}},
		\bauthor{\binits{F.}~\bsnm{Charte}},
		\bauthor{\binits{A.J.}~\bsnm{Rivera}} and
		\bauthor{\binits{M.J.}~\bsnm{del Jesus}},
		\batitle{Choosing the proper autoencoder for feature fusion based on data
			complexity and classifiers: Analysis, tips and guidelines},
		\bjtitle{Information Fusion}
		\bvolume{54}
		(\byear{2020}),
		\bfpage{44}--\blpage{60}.
	\end{barticle}
	\endbibitem
	
	\bibitem{Charte:nonstandard}
	\begin{barticle}
		\bauthor{\binits{D.}~\bsnm{Charte}},
		\bauthor{\binits{F.}~\bsnm{Charte}},
		\bauthor{\binits{S.}~\bsnm{Garc{\'\i}a}} and
		\bauthor{\binits{F.}~\bsnm{Herrera}},
		\batitle{A snapshot on nonstandard supervised learning problems: taxonomy,
			relationships, problem transformations and algorithm adaptations},
		\bjtitle{Progress in Artificial Intelligence}
		\bvolume{8}(\bissue{1})
		(\byear{2019}),
		\bfpage{1}--\blpage{14}.
	\end{barticle}
	\endbibitem
	
	\bibitem{hecht1992theory}
	\begin{bchapter}
		\bauthor{\binits{R.}~\bsnm{Hecht-Nielsen}},
		\bctitle{Theory of the backpropagation neural network},
		in: \bbtitle{Neural networks for perception},
		\bpublisher{Elsevier},
		\byear{1992},
		pp.~\bfpage{65}--\blpage{93}.
	\end{bchapter}
	\endbibitem
	
	\bibitem{robbins1951stochastic}
	\begin{botherref}
		\oauthor{\binits{H.}~\bsnm{Robbins}} and
		\oauthor{\binits{S.}~\bsnm{Monro}},
		A stochastic approximation method,
		\textit{The annals of mathematical statistics}
		(1951),
		400--407.
	\end{botherref}
	\endbibitem
	
	\bibitem{lawrence2000overfitting}
	\begin{bchapter}
		\bauthor{\binits{S.}~\bsnm{Lawrence}} and
		\bauthor{\binits{C.L.}~\bsnm{Giles}},
		\bctitle{Overfitting and neural networks: conjugate gradient and
			backpropagation},
		in: \bbtitle{Proceedings of the IEEE-INNS-ENNS International Joint Conference
			on Neural Networks. IJCNN 2000. Neural Computing: New Challenges and
			Perspectives for the New Millennium},
		Vol.~\bseriesno{1},
		\binstitute{IEEE},
		\byear{2000},
		pp.~\bfpage{114}--\blpage{119}.
	\end{bchapter}
	\endbibitem
	
	\bibitem{ahmadlou2010enhanced}
	\begin{barticle}
		\bauthor{\binits{M.}~\bsnm{Ahmadlou}} and
		\bauthor{\binits{H.}~\bsnm{Adeli}},
		\batitle{Enhanced probabilistic neural network with local decision circles: A
			robust classifier},
		\bjtitle{Integrated Computer-Aided Engineering}
		\bvolume{17}(\bissue{3})
		(\byear{2010}),
		\bfpage{197}--\blpage{210}.
	\end{barticle}
	\endbibitem
	
	\bibitem{benamara2019real}
	\begin{bchapter}
		\bauthor{\binits{N.K.}~\bsnm{Benamara}},
		\bauthor{\binits{M.}~\bsnm{Val-Calvo}},
		\bauthor{\binits{J.R.}~\bsnm{{\'A}lvarez-S{\'a}nchez}},
		\bauthor{\binits{A.}~\bsnm{D{\'\i}az-Morcillo}},
		\bauthor{\binits{J.M.}~\bsnm{Ferr{\'a}ndez-Vicente}},
		\bauthor{\binits{E.}~\bsnm{Fern{\'a}ndez-Jover}} and
		\bauthor{\binits{T.B.}~\bsnm{Stambouli}},
		\bctitle{Real-Time Emotional Recognition for Sociable Robotics Based on Deep
			Neural Networks Ensemble},
		in: \bbtitle{International Work-Conference on the Interplay Between Natural and
			Artificial Computation},
		\binstitute{Springer},
		\byear{2019},
		pp.~\bfpage{171}--\blpage{180}.
	\end{bchapter}
	\endbibitem
	
	\bibitem{PCAHotelling}
	\begin{barticle}
		\bauthor{\binits{H.}~\bsnm{Hotelling}},
		\batitle{Analysis of a complex of statistical variables into principal
			components},
		\bjtitle{Journal of educational psychology}
		\bvolume{24}(\bissue{6})
		(\byear{1933}),
		\bfpage{417}.
	\end{barticle}
	\endbibitem
	
	\bibitem{LDA}
	\begin{barticle}
		\bauthor{\binits{R.A.}~\bsnm{Fisher}},
		\batitle{The statistical utilization of multiple measurements},
		\bjtitle{Annals of Human Genetics}
		\bvolume{8}(\bissue{4})
		(\byear{1938}),
		\bfpage{376}--\blpage{386}.
	\end{barticle}
	\endbibitem
	
	\bibitem{ManifoldLearning}
	\begin{botherref}
		\oauthor{\binits{L.}~\bsnm{Cayton}},
		Algorithms for manifold learning,
		Technical Report,
		University of California at San Diego,
		2005.
	\end{botherref}
	\endbibitem
	
	\bibitem{NonlinearDimRec}
	\begin{bbook}
		\bauthor{\binits{J.A.}~\bsnm{Lee}} and
		\bauthor{\binits{M.}~\bsnm{Verleysen}},
		\bbtitle{Nonlinear dimensionality reduction},
		\bpublisher{Springer Science \& Business Media},
		\byear{2007}.
	\end{bbook}
	\endbibitem
	
	\bibitem{yu2013embedding}
	\begin{bchapter}
		\bauthor{\binits{W.}~\bsnm{Yu}},
		\bauthor{\binits{G.}~\bsnm{Zeng}},
		\bauthor{\binits{P.}~\bsnm{Luo}},
		\bauthor{\binits{F.}~\bsnm{Zhuang}},
		\bauthor{\binits{Q.}~\bsnm{He}} and
		\bauthor{\binits{Z.}~\bsnm{Shi}},
		\bctitle{Embedding with autoencoder regularization},
		in: \bbtitle{Joint European Conference on Machine Learning and Knowledge
			Discovery in Databases},
		\binstitute{Springer},
		\byear{2013},
		pp.~\bfpage{208}--\blpage{223}.
		doi:\doiurl{10.1007/978-3-642-40994-3\_14}.
	\end{bchapter}
	\endbibitem
	
	\bibitem{sakurada}
	\begin{bchapter}
		\bauthor{\binits{M.}~\bsnm{Sakurada}} and
		\bauthor{\binits{T.}~\bsnm{Yairi}},
		\bctitle{Anomaly detection using autoencoders with nonlinear dimensionality
			reduction},
		in: \bbtitle{Proceedings of the MLSDA 2014 2nd Workshop on Machine Learning for
			Sensory Data Analysis},
		\binstitute{ACM},
		\byear{2014},
		pp.~\bfpage{4}--\blpage{11}.
		ISBN \bisbn{978-1-4503-3159-3}.
		doi:\doiurl{10.1145/2689746.2689747}.
	\end{bchapter}
	\endbibitem
	
	\bibitem{park}
	\begin{barticle}
		\bauthor{\binits{S.}~\bsnm{Park}},
		\bauthor{\binits{M.}~\bsnm{Kim}} and
		\bauthor{\binits{S.}~\bsnm{Lee}},
		\batitle{Anomaly Detection for HTTP Using Convolutional Autoencoders},
		\bjtitle{IEEE Access}
		\bvolume{6}
		(\byear{2018}),
		\bfpage{70884}--\blpage{70901}.
		doi:\doiurl{10.1109/ACCESS.2018.2881003}.
	\end{barticle}
	\endbibitem
	
	\bibitem{xie}
	\begin{bchapter}
		\bauthor{\binits{J.}~\bsnm{Xie}},
		\bauthor{\binits{L.}~\bsnm{Xu}} and
		\bauthor{\binits{E.}~\bsnm{Chen}},
		\bctitle{Image denoising and inpainting with deep neural networks},
		in: \bbtitle{Advances in neural information processing systems},
		\byear{2012},
		pp.~\bfpage{341}--\blpage{349}.
	\end{bchapter}
	\endbibitem
	
	\bibitem{speech}
	\begin{bchapter}
		\bauthor{\binits{X.}~\bsnm{Lu}},
		\bauthor{\binits{Y.}~\bsnm{Tsao}},
		\bauthor{\binits{S.}~\bsnm{Matsuda}} and
		\bauthor{\binits{C.}~\bsnm{Hori}},
		\bctitle{Speech enhancement based on deep denoising autoencoder},
		in: \bbtitle{Interspeech},
		\byear{2013},
		pp.~\bfpage{436}--\blpage{440}.
	\end{bchapter}
	\endbibitem
	
	\bibitem{blum2003metaheuristics}
	\begin{barticle}
		\bauthor{\binits{C.}~\bsnm{Blum}} and
		\bauthor{\binits{A.}~\bsnm{Roli}},
		\batitle{Metaheuristics in combinatorial optimization: Overview and conceptual
			comparison},
		\bjtitle{ACM computing surveys (CSUR)}
		\bvolume{35}(\bissue{3})
		(\byear{2003}),
		\bfpage{268}--\blpage{308}.
	\end{barticle}
	\endbibitem
	
	\bibitem{adeli2006cost}
	\begin{bbook}
		\bauthor{\binits{H.}~\bsnm{Adeli}} and
		\bauthor{\binits{K.C.}~\bsnm{Sarma}},
		\bbtitle{Cost optimization of structures: fuzzy logic, genetic algorithms, and
			parallel computing},
		\bpublisher{John Wiley \& Sons},
		\byear{2006}.
	\end{bbook}
	\endbibitem
	
	\bibitem{aarts1997local}
	\begin{botherref}
		\oauthor{\binits{E.}~\bsnm{Aarts}} and
		\oauthor{\binits{J.}~\bsnm{Lenstra}},
		Local Search in Combinatorial Optimization Wiley,
		\textit{New York}
		(1997).
	\end{botherref}
	\endbibitem
	
	\bibitem{den2001design}
	\begin{bchapter}
		\bauthor{\binits{M.}~\bsnm{Den~Besten}},
		\bauthor{\binits{T.}~\bsnm{St{\"u}tzle}} and
		\bauthor{\binits{M.}~\bsnm{Dorigo}},
		\bctitle{Design of iterated local search algorithms},
		in: \bbtitle{Workshops on Applications of Evolutionary Computation},
		\binstitute{Springer},
		\byear{2001},
		pp.~\bfpage{441}--\blpage{451}.
	\end{bchapter}
	\endbibitem
	
	\bibitem{korf1990real}
	\begin{barticle}
		\bauthor{\binits{R.E.}~\bsnm{Korf}},
		\batitle{Real-time heuristic search},
		\bjtitle{Artificial intelligence}
		\bvolume{42}(\bissue{2--3})
		(\byear{1990}),
		\bfpage{189}--\blpage{211}.
	\end{barticle}
	\endbibitem
	
	\bibitem{zhang1999algorithms}
	\begin{bchapter}
		\bauthor{\binits{W.}~\bsnm{Zhang}},
		\bctitle{Algorithms for Combinatorial Optimization},
		in: \bbtitle{State-Space Search},
		\bpublisher{Springer},
		\byear{1999},
		pp.~\bfpage{13}--\blpage{33}.
	\end{bchapter}
	\endbibitem
	
	\bibitem{Neri2008}
	\begin{bchapter}
		\bauthor{\binits{F.}~\bsnm{Neri}},
		\bauthor{\binits{N.}~\bsnm{Kotilainen}} and
		\bauthor{\binits{M.}~\bsnm{Vapa}},
		\bbtitle{A Memetic-Neural Approach to Discover Resources in P2P Networks},
		in: \bbtitle{Recent Advances in Evolutionary Computation for Combinatorial
			Optimization},
		\beditor{\binits{C.}~\bsnm{Cotta}} and
		\beditor{\binits{J.}~\bsnm{van Hemert}}, eds,
		\bpublisher{Springer Berlin Heidelberg},
		\blocation{Berlin, Heidelberg},
		\byear{2008},
		pp.~\bfpage{113}--\blpage{129}.
		ISBN \bisbn{978-3-540-70807-0}.
		doi:\doiurl{10.1007/978-3-540-70807-0\_8}.
	\end{bchapter}
	\endbibitem
	
	\bibitem{van1987simulated}
	\begin{bchapter}
		\bauthor{\binits{P.J.}~\bsnm{Van~Laarhoven}} and
		\bauthor{\binits{E.H.}~\bsnm{Aarts}},
		\bctitle{Simulated annealing},
		in: \bbtitle{Simulated annealing: Theory and applications},
		\bpublisher{Springer},
		\byear{1987},
		pp.~\bfpage{7}--\blpage{15}.
	\end{bchapter}
	\endbibitem
	
	\bibitem{battiti1994reactive}
	\begin{barticle}
		\bauthor{\binits{R.}~\bsnm{Battiti}} and
		\bauthor{\binits{G.}~\bsnm{Tecchiolli}},
		\batitle{The reactive tabu search},
		\bjtitle{ORSA journal on computing}
		\bvolume{6}(\bissue{2})
		(\byear{1994}),
		\bfpage{126}--\blpage{140}.
	\end{barticle}
	\endbibitem
	
	\bibitem{freitas2009review}
	\begin{bchapter}
		\bauthor{\binits{A.A.}~\bsnm{Freitas}},
		\bctitle{A review of evolutionary algorithms for data mining},
		in: \bbtitle{Data Mining and Knowledge Discovery Handbook},
		\bpublisher{Springer},
		\byear{2009},
		pp.~\bfpage{371}--\blpage{400}.
	\end{bchapter}
	\endbibitem
	
	\bibitem{Bck1993AnOO}
	\begin{barticle}
		\bauthor{\binits{T.}~\bsnm{B{\"a}ck}} and
		\bauthor{\binits{H.-P.}~\bsnm{Schwefel}},
		\batitle{An Overview of Evolutionary Algorithms for Parameter Optimization},
		\bjtitle{Evolutionary Computation}
		\bvolume{1}
		(\byear{1993}),
		\bfpage{1}--\blpage{23}.
	\end{barticle}
	\endbibitem
	
	\bibitem{wang2019optimizing}
	\begin{barticle}
		\bauthor{\binits{Q.}~\bsnm{Wang}},
		\bauthor{\binits{H.-L.}~\bsnm{Liu}},
		\bauthor{\binits{J.}~\bsnm{Yuan}} and
		\bauthor{\binits{L.}~\bsnm{Chen}},
		\batitle{Optimizing the energy-spectrum efficiency of cellular systems by
			evolutionary multi-objective algorithm},
		\bjtitle{Integrated Computer-Aided Engineering}
		\bvolume{26}(\bissue{2})
		(\byear{2019}),
		\bfpage{207}--\blpage{220}.
	\end{barticle}
	\endbibitem
	
	\bibitem{kyriklidis2016evolutionary}
	\begin{barticle}
		\bauthor{\binits{C.}~\bsnm{Kyriklidis}} and
		\bauthor{\binits{G.}~\bsnm{Dounias}},
		\batitle{Evolutionary computation for resource leveling optimization in project
			management},
		\bjtitle{Integrated Computer-Aided Engineering}
		\bvolume{23}(\bissue{2})
		(\byear{2016}),
		\bfpage{173}--\blpage{184}.
	\end{barticle}
	\endbibitem
	
	\bibitem{kociecki2014two}
	\begin{barticle}
		\bauthor{\binits{M.}~\bsnm{Kociecki}} and
		\bauthor{\binits{H.}~\bsnm{Adeli}},
		\batitle{Two-phase genetic algorithm for topology optimization of free-form
			steel space-frame roof structures with complex curvatures},
		\bjtitle{Engineering Applications of Artificial Intelligence}
		\bvolume{32}
		(\byear{2014}),
		\bfpage{218}--\blpage{227}.
	\end{barticle}
	\endbibitem
	
	\bibitem{kociecki2015shape}
	\begin{barticle}
		\bauthor{\binits{M.}~\bsnm{Kociecki}} and
		\bauthor{\binits{H.}~\bsnm{Adeli}},
		\batitle{Shape optimization of free-form steel space-frame roof structures with
			complex geometries using evolutionary computing},
		\bjtitle{Engineering Applications of Artificial Intelligence}
		\bvolume{38}
		(\byear{2015}),
		\bfpage{168}--\blpage{182}.
	\end{barticle}
	\endbibitem
	
	\bibitem{blum2002ant}
	\begin{bchapter}
		\bauthor{\binits{C.}~\bsnm{Blum}},
		\bctitle{Ant colony optimization for the edge-weighted k-cardinality tree
			problem},
		in: \bbtitle{Proceedings of the 4th Annual Conference on Genetic and
			Evolutionary Computation},
		\binstitute{Morgan Kaufmann Publishers Inc.},
		\byear{2002},
		pp.~\bfpage{27}--\blpage{34}.
	\end{bchapter}
	\endbibitem
	
	\bibitem{poli2007particle}
	\begin{barticle}
		\bauthor{\binits{R.}~\bsnm{Poli}},
		\bauthor{\binits{J.}~\bsnm{Kennedy}} and
		\bauthor{\binits{T.}~\bsnm{Blackwell}},
		\batitle{Particle swarm optimization},
		\bjtitle{Swarm intelligence}
		\bvolume{1}(\bissue{1})
		(\byear{2007}),
		\bfpage{33}--\blpage{57}.
	\end{barticle}
	\endbibitem
	
	\bibitem{foster2001computational}
	\begin{barticle}
		\bauthor{\binits{J.A.}~\bsnm{Foster}},
		\batitle{Computational genetics: Evolutionary computation},
		\bjtitle{Nature Reviews Genetics}
		\bvolume{2}(\bissue{6})
		(\byear{2001}),
		\bfpage{428}.
	\end{barticle}
	\endbibitem
	
	\bibitem{back2018evolutionary}
	\begin{bbook}
		\bauthor{\binits{T.}~\bsnm{B{\"a}ck}},
		\bauthor{\binits{D.B.}~\bsnm{Fogel}} and
		\bauthor{\binits{Z.}~\bsnm{Michalewicz}},
		\bbtitle{Evolutionary computation 1: Basic algorithms and operators},
		\bpublisher{CRC press},
		\byear{2018}.
	\end{bbook}
	\endbibitem
	
	\bibitem{davis1991handbook}
	\begin{botherref}
		\oauthor{\binits{L.}~\bsnm{Davis}},
		Handbook of genetic algorithms
		(1991).
	\end{botherref}
	\endbibitem
	
	\bibitem{storn1997differential}
	\begin{barticle}
		\bauthor{\binits{R.}~\bsnm{Storn}} and
		\bauthor{\binits{K.}~\bsnm{Price}},
		\batitle{Differential evolution--a simple and efficient heuristic for global
			optimization over continuous spaces},
		\bjtitle{Journal of global optimization}
		\bvolume{11}(\bissue{4})
		(\byear{1997}),
		\bfpage{341}--\blpage{359}.
	\end{barticle}
	\endbibitem
	
	\bibitem{rechenberg1984evolution}
	\begin{bchapter}
		\bauthor{\binits{I.}~\bsnm{Rechenberg}},
		\bctitle{The evolution strategy. a mathematical model of darwinian evolution},
		in: \bbtitle{Synergetics—from microscopic to macroscopic order},
		\bpublisher{Springer},
		\byear{1984},
		pp.~\bfpage{122}--\blpage{132}.
	\end{bchapter}
	\endbibitem
	
	\bibitem{kim2001discrete}
	\begin{barticle}
		\bauthor{\binits{H.}~\bsnm{KIM}} and
		\bauthor{\binits{H.}~\bsnm{ADELI}},
		\batitle{Discrete cost optimization of composite floors using a floating-point
			genetic algorithm},
		\bjtitle{Engineering Optimization}
		\bvolume{33}(\bissue{4})
		(\byear{2001}),
		\bfpage{485}--\blpage{501}.
	\end{barticle}
	\endbibitem
	
	\bibitem{friedrichs2005evolutionary}
	\begin{barticle}
		\bauthor{\binits{F.}~\bsnm{Friedrichs}} and
		\bauthor{\binits{C.}~\bsnm{Igel}},
		\batitle{Evolutionary tuning of multiple SVM parameters},
		\bjtitle{Neurocomputing}
		\bvolume{64}
		(\byear{2005}),
		\bfpage{107}--\blpage{117}.
	\end{barticle}
	\endbibitem
	
	\bibitem{young2015optimizing}
	\begin{bchapter}
		\bauthor{\binits{S.R.}~\bsnm{Young}},
		\bauthor{\binits{D.C.}~\bsnm{Rose}},
		\bauthor{\binits{T.P.}~\bsnm{Karnowski}},
		\bauthor{\binits{S.-H.}~\bsnm{Lim}} and
		\bauthor{\binits{R.M.}~\bsnm{Patton}},
		\bctitle{Optimizing deep learning hyper-parameters through an evolutionary
			algorithm},
		in: \bbtitle{Proceedings of the Workshop on Machine Learning in
			High-Performance Computing Environments},
		\binstitute{ACM},
		\byear{2015},
		p.~\bfpage{4}.
	\end{bchapter}
	\endbibitem
	
	\bibitem{ANNEvo1}
	\begin{bchapter}
		\bauthor{\binits{H.}~\bsnm{Kitano}},
		\bctitle{Empirical Studies on the Speed of Convergence of Neural Network
			Training Using Genetic Algorithms.},
		in: \bbtitle{AAAI},
		\byear{1990},
		pp.~\bfpage{789}--\blpage{795}.
	\end{bchapter}
	\endbibitem
	
	\bibitem{ANNEvo2}
	\begin{bchapter}
		\bauthor{\binits{M.}~\bsnm{Scholz}},
		\bctitle{A learning strategy for neural networks based on a modified
			evolutionary strategy},
		in: \bbtitle{International Conference on Parallel Problem Solving from Nature},
		\binstitute{Springer},
		\byear{1990},
		pp.~\bfpage{314}--\blpage{318}.
	\end{bchapter}
	\endbibitem
	
	\bibitem{ANNEvo3}
	\begin{bchapter}
		\bauthor{\binits{L.D.}~\bsnm{Whitley}} and
		\bauthor{\binits{T.}~\bsnm{Hanson}},
		\bctitle{Optimizing Neural Networks Using FasterMore Accurate Genetic Search},
		in: \bbtitle{Proceedings of the 3rd international conference on genetic
			algorithms},
		\binstitute{Morgan Kaufmann Publishers Inc.},
		\byear{1989},
		pp.~\bfpage{391}--\blpage{397}.
	\end{bchapter}
	\endbibitem
	
	\bibitem{ANNEvo4}
	\begin{barticle}
		\bauthor{\binits{K.}~\bsnm{Chellapilla}} and
		\bauthor{\binits{D.B.}~\bsnm{Fogel}},
		\batitle{Evolving neural networks to play checkers without relying on expert
			knowledge},
		\bjtitle{IEEE transactions on neural networks}
		\bvolume{10}(\bissue{6})
		(\byear{1999}),
		\bfpage{1382}--\blpage{1391}.
	\end{barticle}
	\endbibitem
	
	\bibitem{floreano2008neuroevolution}
	\begin{barticle}
		\bauthor{\binits{D.}~\bsnm{Floreano}},
		\bauthor{\binits{P.}~\bsnm{D{\"u}rr}} and
		\bauthor{\binits{C.}~\bsnm{Mattiussi}},
		\batitle{Neuroevolution: from architectures to learning},
		\bjtitle{Evolutionary Intelligence}
		\bvolume{1}(\bissue{1})
		(\byear{2008}),
		\bfpage{47}--\blpage{62}.
	\end{barticle}
	\endbibitem
	
	\bibitem{gruau1994automatic}
	\begin{barticle}
		\bauthor{\binits{F.}~\bsnm{Gruau}},
		\batitle{Automatic definition of modular neural networks},
		\bjtitle{Adaptive behavior}
		\bvolume{3}(\bissue{2})
		(\byear{1994}),
		\bfpage{151}--\blpage{183}.
	\end{barticle}
	\endbibitem
	
	\bibitem{stanley2019designing}
	\begin{barticle}
		\bauthor{\binits{K.O.}~\bsnm{Stanley}},
		\bauthor{\binits{J.}~\bsnm{Clune}},
		\bauthor{\binits{J.}~\bsnm{Lehman}} and
		\bauthor{\binits{R.}~\bsnm{Miikkulainen}},
		\batitle{Designing neural networks through neuroevolution},
		\bjtitle{Nature Machine Intelligence}
		\bvolume{1}(\bissue{1})
		(\byear{2019}),
		\bfpage{24}--\blpage{35}.
	\end{barticle}
	\endbibitem
	
	\bibitem{AutoML}
	\begin{bchapter}
		\bauthor{\binits{I.}~\bsnm{Guyon}},
		\bauthor{\binits{L.}~\bsnm{Sun-Hosoya}},
		\bauthor{\binits{M.}~\bsnm{Boull{\'e}}},
		\bauthor{\binits{H.J.}~\bsnm{Escalante}},
		\bauthor{\binits{S.}~\bsnm{Escalera}},
		\bauthor{\binits{Z.}~\bsnm{Liu}},
		\bauthor{\binits{D.}~\bsnm{Jajetic}},
		\bauthor{\binits{B.}~\bsnm{Ray}},
		\bauthor{\binits{M.}~\bsnm{Saeed}},
		\bauthor{\binits{M.}~\bsnm{Sebag}} \betal,
		\bctitle{Analysis of the AutoML Challenge Series 2015--2018},
		in: \bbtitle{Automated Machine Learning},
		\bpublisher{Springer},
		\byear{2019},
		pp.~\bfpage{177}--\blpage{219}.
	\end{bchapter}
	\endbibitem
	
	\bibitem{ying2019bench}
	\begin{bchapter}
		\bauthor{\binits{C.}~\bsnm{Ying}},
		\bauthor{\binits{A.}~\bsnm{Klein}},
		\bauthor{\binits{E.}~\bsnm{Christiansen}},
		\bauthor{\binits{E.}~\bsnm{Real}},
		\bauthor{\binits{K.}~\bsnm{Murphy}} and
		\bauthor{\binits{F.}~\bsnm{Hutter}},
		\bctitle{NAS-Bench-101: Towards Reproducible Neural Architecture Search},
		in: \bbtitle{International Conference on Machine Learning},
		\byear{2019},
		pp.~\bfpage{7105}--\blpage{7114}.
	\end{bchapter}
	\endbibitem
	
	\bibitem{feurer2015efficient}
	\begin{bchapter}
		\bauthor{\binits{M.}~\bsnm{Feurer}},
		\bauthor{\binits{A.}~\bsnm{Klein}},
		\bauthor{\binits{K.}~\bsnm{Eggensperger}},
		\bauthor{\binits{J.}~\bsnm{Springenberg}},
		\bauthor{\binits{M.}~\bsnm{Blum}} and
		\bauthor{\binits{F.}~\bsnm{Hutter}},
		\bctitle{Efficient and robust automated machine learning},
		in: \bbtitle{Advances in neural information processing systems},
		\byear{2015},
		pp.~\bfpage{2962}--\blpage{2970}.
	\end{bchapter}
	\endbibitem
	
	\bibitem{van2019automatic}
	\begin{bchapter}
		\bauthor{\binits{B.}~\bsnm{van Stein}},
		\bauthor{\binits{H.}~\bsnm{Wang}} and
		\bauthor{\binits{T.}~\bsnm{B{\"a}ck}},
		\bctitle{Automatic Configuration of Deep Neural Networks with Parallel
			Efficient Global Optimization},
		in: \bbtitle{2019 International Joint Conference on Neural Networks (IJCNN)},
		\binstitute{IEEE},
		\byear{2019},
		pp.~\bfpage{1}--\blpage{7}.
	\end{bchapter}
	\endbibitem
	
	\bibitem{Jin2018AutoKerasAE}
	\begin{botherref}
		\oauthor{\binits{H.}~\bsnm{Jin}},
		\oauthor{\binits{Q.}~\bsnm{Song}} and
		\oauthor{\binits{X.}~\bsnm{Hu}},
		Auto-Keras: An Efficient Neural Architecture Search System,
		in: \textit{Proceedings of the 25th ACM SIGKDD International Conference on
			Knowledge Discovery \& Data Mining},
		ACM], pages={1946–-1956}, year={2019},.
	\end{botherref}
	\endbibitem
	
	\bibitem{price2006differential}
	\begin{bbook}
		\bauthor{\binits{K.}~\bsnm{Price}},
		\bauthor{\binits{R.M.}~\bsnm{Storn}} and
		\bauthor{\binits{J.A.}~\bsnm{Lampinen}},
		\bbtitle{Differential evolution: a practical approach to global optimization},
		\bpublisher{Springer Science \& Business Media},
		\byear{2006}.
	\end{bbook}
	\endbibitem
	
	\bibitem{cifar10}
	\begin{botherref}
		\oauthor{\binits{A.}~\bsnm{Krizhevsky}},
		Learning Multiple Layers of Features from Tiny Images
		(2009).
	\end{botherref}
	\endbibitem
	
	\bibitem{delicious}
	\begin{bchapter}
		\bauthor{\binits{G.}~\bsnm{Tsoumakas}},
		\bauthor{\binits{I.}~\bsnm{Katakis}} and
		\bauthor{\binits{I.}~\bsnm{Vlahavas}},
		\bctitle{Effective and Efficient Multilabel Classification in Domains with
			Large Number of Labels},
		in: \bbtitle{Proc. ECML/PKDD Workshop on Mining Multidimensional Data, Antwerp,
			Belgium, MMD08},
		\byear{2008},
		pp.~\bfpage{30}--\blpage{44}.
	\end{bchapter}
	\endbibitem
	
	\bibitem{fashion}
	\begin{botherref}
		\oauthor{\binits{H.}~\bsnm{Xiao}},
		\oauthor{\binits{K.}~\bsnm{Rasul}} and
		\oauthor{\binits{R.}~\bsnm{Vollgraf}},
		Fashion-MNIST: a Novel Image Dataset for Benchmarking Machine Learning
		Algorithms,
		2017.
	\end{botherref}
	\endbibitem
	
	\bibitem{glass}
	\begin{botherref}
		\oauthor{\binits{I.W.}~\bsnm{Evett}} and
		\oauthor{\binits{E.J.}~\bsnm{Spiehler}},
		Rule induction in forensic science,
		\textit{KBS in Goverment}
		(1987),
		107--118.
	\end{botherref}
	\endbibitem
	
	\bibitem{ionosphere}
	\begin{barticle}
		\bauthor{\binits{V.G.}~\bsnm{Sigillito}},
		\bauthor{\binits{S.P.}~\bsnm{Wing}},
		\bauthor{\binits{L.V.}~\bsnm{Hutton}} and
		\bauthor{\binits{K.B.}~\bsnm{Baker}},
		\batitle{Classification of radar returns from the ionosphere using neural
			networks},
		\bjtitle{Johns Hopkins APL Technical Digest}
		\bvolume{10}(\bissue{3})
		(\byear{1989}),
		\bfpage{262}--\blpage{266}.
	\end{barticle}
	\endbibitem
	
	\bibitem{mnist}
	\begin{barticle}
		\bauthor{\binits{L.}~\bsnm{Deng}},
		\batitle{The MNIST database of handwritten digit images for machine learning
			research [best of the web]},
		\bjtitle{IEEE Signal Processing Magazine}
		\bvolume{29}(\bissue{6})
		(\byear{2012}),
		\bfpage{141}--\blpage{142}.
	\end{barticle}
	\endbibitem
	
	\bibitem{semeion}
	\begin{barticle}
		\bauthor{\binits{M.}~\bsnm{Buscema}},
		\batitle{Metanet*: The theory of independent judges},
		\bjtitle{Substance use \& misuse}
		\bvolume{33}(\bissue{2})
		(\byear{1998}),
		\bfpage{439}--\blpage{461}.
	\end{barticle}
	\endbibitem
	
	\bibitem{sonar}
	\begin{barticle}
		\bauthor{\binits{R.P.}~\bsnm{Gorman}} and
		\bauthor{\binits{T.J.}~\bsnm{Sejnowski}},
		\batitle{Analysis of hidden units in a layered network trained to classify
			sonar targets},
		\bjtitle{Neural networks}
		\bvolume{1}(\bissue{1})
		(\byear{1988}),
		\bfpage{75}--\blpage{89}.
	\end{barticle}
	\endbibitem
	
	\bibitem{spect}
	\begin{barticle}
		\bauthor{\binits{L.A.}~\bsnm{Kurgan}},
		\bauthor{\binits{K.J.}~\bsnm{Cios}},
		\bauthor{\binits{R.}~\bsnm{Tadeusiewicz}},
		\bauthor{\binits{M.}~\bsnm{Ogiela}} and
		\bauthor{\binits{L.S.}~\bsnm{Goodenday}},
		\batitle{Knowledge discovery approach to automated cardiac SPECT diagnosis},
		\bjtitle{Artificial intelligence in medicine}
		\bvolume{23}(\bissue{2})
		(\byear{2001}),
		\bfpage{149}--\blpage{169}.
	\end{barticle}
	\endbibitem
	
	\bibitem{ArchLinux}
	\begin{bchapter}
		\bauthor{\binits{J.D.}~\bsnm{Castro}},
		\bctitle{Arch linux},
		in: \bbtitle{Introducing Linux Distros},
		\bpublisher{Springer},
		\byear{2016},
		pp.~\bfpage{235}--\blpage{252}.
	\end{bchapter}
	\endbibitem
	
	\bibitem{cuda}
	\begin{barticle}
		\bauthor{\binits{M.}~\bsnm{Garland}},
		\bauthor{\binits{S.}~\bsnm{Le~Grand}},
		\bauthor{\binits{J.}~\bsnm{Nickolls}},
		\bauthor{\binits{J.}~\bsnm{Anderson}},
		\bauthor{\binits{J.}~\bsnm{Hardwick}},
		\bauthor{\binits{S.}~\bsnm{Morton}},
		\bauthor{\binits{E.}~\bsnm{Phillips}},
		\bauthor{\binits{Y.}~\bsnm{Zhang}} and
		\bauthor{\binits{V.}~\bsnm{Volkov}},
		\batitle{Parallel computing experiences with CUDA},
		\bjtitle{IEEE micro}
		\bvolume{28}(\bissue{4})
		(\byear{2008}),
		\bfpage{13}--\blpage{27}.
	\end{barticle}
	\endbibitem
	
	\bibitem{cudnn}
	\begin{botherref}
		\oauthor{\binits{S.}~\bsnm{Chetlur}},
		\oauthor{\binits{C.}~\bsnm{Woolley}},
		\oauthor{\binits{P.}~\bsnm{Vandermersch}},
		\oauthor{\binits{J.}~\bsnm{Cohen}},
		\oauthor{\binits{J.}~\bsnm{Tran}},
		\oauthor{\binits{B.}~\bsnm{Catanzaro}} and
		\oauthor{\binits{E.}~\bsnm{Shelhamer}},
		cudnn: Efficient primitives for deep learning,
		\textit{arXiv preprint arXiv:1410.0759}
		(2014).
	\end{botherref}
	\endbibitem
	
	\bibitem{tensorflow}
	\begin{bchapter}
		\bauthor{\binits{M.}~\bsnm{Abadi}},
		\bauthor{\binits{P.}~\bsnm{Barham}},
		\bauthor{\binits{J.}~\bsnm{Chen}},
		\bauthor{\binits{Z.}~\bsnm{Chen}},
		\bauthor{\binits{A.}~\bsnm{Davis}},
		\bauthor{\binits{J.}~\bsnm{Dean}},
		\bauthor{\binits{M.}~\bsnm{Devin}},
		\bauthor{\binits{S.}~\bsnm{Ghemawat}},
		\bauthor{\binits{G.}~\bsnm{Irving}},
		\bauthor{\binits{M.}~\bsnm{Isard}} \betal,
		\bctitle{Tensorflow: A system for large-scale machine learning},
		in: \bbtitle{12th $\{$USENIX$\}$ Symposium on Operating Systems Design and
			Implementation ($\{$OSDI$\}$ 16)},
		\byear{2016},
		pp.~\bfpage{265}--\blpage{283}.
	\end{bchapter}
	\endbibitem
	
	\bibitem{keras}
	\begin{barticle}
		\bauthor{\binits{F.}~\bsnm{Chollet}} \betal,
		\batitle{Keras: Deep learning library for theano and tensorflow},
		\bjtitle{URL: https://keras. io/k}
		\bvolume{7}(\bissue{8})
		(\byear{2015}),
		\bfpage{T1}.
	\end{barticle}
	\endbibitem
	
	\bibitem{Charte:rutaPackage}
	\begin{barticle}
		\bauthor{\binits{D.}~\bsnm{Charte}},
		\bauthor{\binits{F.}~\bsnm{Herrera}} and
		\bauthor{\binits{F.}~\bsnm{Charte}},
		\batitle{{Ruta: implementations of neural autoencoders in R}},
		\bjtitle{Knowledge-Based Systems}
		\bvolume{174}
		(\byear{2019}),
		\bfpage{4}--\blpage{8}.
	\end{barticle}
	\endbibitem
	
	\bibitem{RMSProp}
	\begin{barticle}
		\bauthor{\binits{T.}~\bsnm{Tieleman}} and
		\bauthor{\binits{G.}~\bsnm{Hinton}},
		\batitle{Lecture 6.5-rmsprop: Divide the gradient by a running average of its
			recent magnitude},
		\bjtitle{COURSERA: Neural networks for machine learning}
		\bvolume{4}(\bissue{2})
		(\byear{2012}),
		\bfpage{26}--\blpage{31}.
	\end{barticle}
	\endbibitem
	
\end{thebibliography}

%

\end{document}